\documentclass[onecolumn, amsmath, amssymb, floatfix, superscriptaddress]{revtex4-2}

\usepackage{graphicx}       %
\usepackage{dcolumn}        %
\usepackage{bm}             %
\usepackage{xcolor}         %
\usepackage{amsmath , booktabs}
\usepackage[T1]{fontenc}
\usepackage{multirow}
\usepackage[ruled,vlined]{algorithm2e} %
\usepackage{hyperref}       %
\hypersetup{
  colorlinks = true,
  linkcolor  = blue,
  citecolor  = blue,
  urlcolor   = blue
}

\newcommand{\sindy}{\textsc{SINDy}}
\newcommand{\esindy}{\textsc{E-SINDy}}
\newcommand{\autosindy}{\textsc{AutoSINDy}}
\newcommand{\pysr}{\textsc{PySR}}
\newcommand{\pysindy}{\textsc{PySINDy}}
\colorlet{revcolor}{black}
\newcommand{\rev}[1]{\textcolor{revcolor}{#1}}

\DeclareMathOperator*{\argmin}{arg\,min}

\begin{document}

\title{\rev{\autosindy:} Discovery of Nonlinear Dynamics with Automated \rev{Library Construction}}

\author{Mohammad Amin Basiri}
\email{ma.basiri@ou.edu}
\affiliation{Data Science and Analytics Institute, University of Oklahoma, Norman, OK, USA}

\author{Charles Nicholson}
\email{cnicholson@ou.edu}
\affiliation{Data Science and Analytics Institute, University of Oklahoma, Norman, OK, USA}
\affiliation{School of Industrial and Systems Engineering, University of Oklahoma, Norman, OK, USA}

\date{\today}

\begin{abstract}
Discovering governing equations from observational data remains a fundamental challenge in scientific modeling, particularly when the underlying mathematical structure is unknown.  Traditional sparse identification methods like \sindy{} excel at discovering parsimonious models but require researchers to specify candidate basis functions a priori, a limitation that often leads to model failure when critical terms are omitted or when systems exhibit unconventional dynamics.  Purely symbolic regression approaches offer unlimited flexibility but struggle with noise sensitivity and frequently produce overly complex, unstable equations.  We present \autosindy, a hybrid Discovery-then-Solve framework that combines the exploratory power of symbolic regression with the robust sparsity-promoting capabilities of \sindy.  Our method operates in three stages:
(1)~\pysr-based symbolic regression discovers candidate functional forms from \rev{random subsets of the data}; (2)~a curation pipeline decomposes, expands, and filters these expressions using collinearity analysis to construct a minimal yet comprehensive library; and (3)~\rev{the weak form of} \sindy{} identifies sparse governing equations from this custom-tailored library.  Extensive experiments across \rev{six} canonical nonlinear systems (oscillatory, chaotic, etc.)\rev{, with noise of up to 20\% of each variable's standard deviation,} demonstrate that \autosindy{} \rev{achieves} a recovery rate \rev{(new-trajectory $R^2 > 0.99$)} of \rev{89.7}\% across all trials\rev{, against at most 45.0\% for the baselines, and 67.5\% at the highest noise level}.  Compared with standard \sindy{} using enriched libraries\rev{, its weak form,} and standalone symbolic regression, \autosindy{} achieves higher predictive accuracy, superior generalization to unseen trajectories, and \rev{equations close to the ground-truth complexity}.  These results show that \autosindy{} successfully identifies correct governing terms without prior domain knowledge, effectively automating the feature-engineering bottleneck in system identification and offering a pathway for discovering complex dynamics in biological and physical systems where the mathematical form is unknown.
\end{abstract}


\maketitle

\section{\label{sec:intro}Introduction}

The discovery of governing mathematical equations from observational data is one of the deepest aspirations of the scientific method.  From Johannes Kepler, who inferred the laws of planetary motion from decades of astronomical records, to Newton formulating the classical mathematical description of universal gravitation and the laws of motion~\cite{campsvalls2023discoveringcausalrelationsequations}, scientists have long sought compact, interpretable models that reveal the mechanistic structure beneath experimental observations.  Today, this ambition is being transformed and accelerated by the rise of scientific machine learning.  Artificial intelligence and data-driven modeling are now deeply integrated into the entire scientific workflow, from automated hypothesis generation and experimental design to the \rev{data-driven modeling of complex dynamical systems} directly from time-series data~\cite{Wang2023,Kutz2025}.  The governing equations of dynamical systems occupy a privileged position in this landscape: they are the ultimate compressed representation of a physical or biological process, encoding causality, generalizability, and mechanistic interpretability in a form that no black-box model can provide.  Recovering these equations from data has profound implications across domains where first-principles derivations are intractable, such as fluid mechanics~\cite{Callaham_Brunton_Loiseau_2022}, plasma physics~\cite{plasma}, climate models~\cite{climate}, neuroscience~\cite{brenner2022tractabledendriticrnnsreconstructing}, and systems biology~\cite{Metayer2025}.  As both the volume of scientific data and the computational power available for analysis continue to grow, the need for principled, automated, and assumption-free methods for equation discovery has never been more urgent.

Two dominant and complementary paradigms have emerged for data-driven discovery of governing equations: sparse regression-based identification and symbolic regression.  The Sparse Identification of Nonlinear Dynamics (\sindy) framework, introduced by Brunton, Proctor, and Kutz~\cite{Brunton2016}, formulates equation discovery as a sparse regression problem over a user-specified library of candidate functions.  Assuming that most dynamical systems can be described by only a few active terms from a large dictionary of possible nonlinearities, \sindy{} applies sequentially thresholded least squares (STLSQ) to identify the minimal subset of library functions that best fits the measured time derivatives of the state. This elegant formulation produces interpretable, parsimonious models and requires substantially less data than neural-network-based approaches.  Its generalization to partial differential equations (PDEs) via PDE-FIND~\cite{Rudy2017} further broadened its scope, and \rev{a rich family of extensions followed}, including \sindy{} with control inputs~\cite{BruntonControl2016}, implicit formulations for rational nonlinearities~\cite{Kaheman2020}, and autoencoder-based coordinate discovery~\cite{Champion2019}\rev{, many of them available in the open-source \pysindy{} package~\cite{deSilva2020,Kaptanoglu2022}}.  Subsequent innovations have markedly improved robustness and model selection\rev{.} Sparse Relaxed Regularized Regression (SR3)~\cite{9194760} \rev{relaxes the sparse regression with an auxiliary variable, which also accommodates physical constraints and outliers.} \rev{W}eak-form \sindy~\rev{\cite{MessengerODE2021,Messenger2021}} bypasses noisy derivative estimation by integrating against test functions\rev{.} \sindy-AIC~\cite{Mangan2017} applies the Akaike information criterion (AIC) to automatically select \rev{the model with the best balance of fit and complexity.} \esindy~\cite{Fasel2022} leverages bootstrap aggregating (bagging) to identify probabilistic models robust to limited data and high noise.  \rev{More recent Bayesian variants quantify the uncertainty of the identified terms and coefficients~\cite{Zhang2018,Nayek2021,MartinaPerez2021,Hirsh2022,Niven2024,Fung2025}, and other variants estimate the noise-free states together with the model~\cite{Kaheman2022,Ribera2022,Lejarza2022,Hokanson2023}.}  In parallel, symbolic regression offers a fundamentally different and complementary approach: rather than selecting terms from a pre-built dictionary, it searches the space of mathematical expressions using evolutionary algorithms.  Following the foundational work of Koza~\cite{Koza1992} and the landmark \rev{work} of Schmidt and Lipson~\cite{Schmidt2009}, modern tools such as \pysr~\cite{Cranmer2023} have established symbolic regression as a practical method for recovering governing equations across physical, biological, and epidemiological systems~\cite{Brum2025}.  \rev{Many other symbolic regression tools are in use~\cite{Udrescu2020,Guimera2020,Bartlett2024,Burlacu2020,Aldeia2026,Cornelio2023,CoryWright2024}, and the SRBench benchmark compares many of them~\cite{LaCava2021}.}

However, despite these significant advances, critical limitations
prevent either paradigm from serving as a fully general-purpose
solution for automated equation discovery.  The central bottleneck of
\sindy-based methods is their dependence on a user-specified candidate
library: the correct governing terms must be anticipated and explicitly
included prior to any regression.  If a critical basis function is
omitted, whether due to unconventional dynamics, unexpected
nonlinearities, or genuine domain ignorance, the method fails to
recover the true model regardless of the downstream
optimizer~\rev{\cite{Brunton2016,Metayer2025}}.  Enriching the library with an
exhaustive set of candidate terms partially addresses omission errors
but introduces severe multicollinearity among features, \rev{which inflates
coefficient uncertainty and destabilizes sparse
regression~\cite{Fasel2022,Metayer2025}}.  A major remaining challenge
is therefore to construct libraries that are simultaneously expressive
enough to contain the true governing terms and compact enough to avoid
redundancy and collinearity, a requirement that cannot be satisfied by
any fixed or generic library in the absence of domain knowledge.
Symbolic regression~\rev{\cite{Cranmer2023}} addresses the library problem directly, but this
flexibility comes at a steep price\rev{. The} combinatorial nature of
expression tree search renders it computationally
intensive~\rev{\cite{virgolin2022symbolicregressionnphard,Brum2025}}, highly
sensitive to measurement noise, and prone to producing
expressions that are numerically unstable when integrated as
differential equations.  Moreover,
because symbolic regression methods optimize each state variable
independently, they struggle to enforce the coupled structure of
multi-dimensional dynamical systems~\cite{Metayer2025}.  The broader challenge, which is automatically constructing a minimal, physics-appropriate function library from raw data without any prior knowledge of the system's mathematical form, remains an open and critical problem in scientific machine learning.

\begin{figure}[t]
	\centering
	\includegraphics[width=0.85\columnwidth]{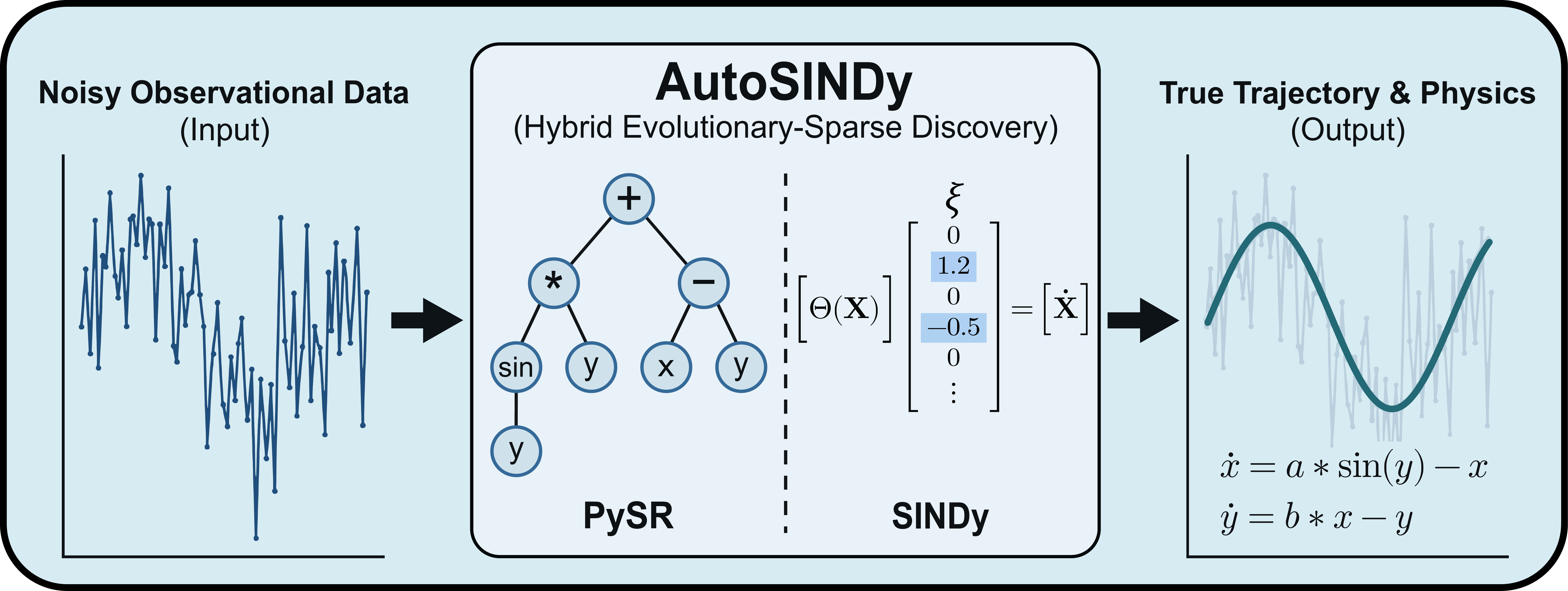}
	\caption{Conceptual overview of \autosindy.  Noisy observational
    data (left) enters a hybrid evolutionary-sparse discovery engine
    (centre): \pysr{} searches the symbolic space via expression-tree
    evolution, producing a set of candidate functional forms that is
    then filtered by \sindy's sparse regression.  The output (right)
    is a concise, closed-form governing equation that faithfully
    reconstructs the underlying physics. \rev{Designed and drawn by the
    authors in Inkscape. Some icons and vector elements were drafted with
    Claude (Opus 5.5, Anthropic) and refined by the authors.}}
	\label{fig:concept}
\end{figure}

In this work, we present \autosindy, a hybrid
\emph{Discovery-then-Solve} framework designed to directly bridge this
gap.  The key insight is a structural decomposition of the equation
discovery problem: symbolic regression, applied \rev{to small random
subsets of the data}, need not solve the complete discovery
problem, it needs only surface plausible candidate functional forms
efficiently.  These candidates are then systematically curated and
handed to \sindy{} for precise, sparse coefficient identification.
This separation of \emph{structure discovery} (what functional forms
might appear) from \emph{sparse selection} (which terms are active and
with what coefficients) allows each component to operate in the regime
where it excels.  \autosindy{} thus recovers the expressive power of
symbolic regression while retaining the interpretability, efficiency,
and robustness of \sindy{}, without requiring any domain knowledge
about the system's mathematical form. Figure~\ref{fig:concept} provides a high-level illustration of this pipeline.

The primary contributions of this work are as follows.
\begin{enumerate}
  \item \textbf{Automated, \rev{data-driven} library generation.}
        We propose a \pysr-based discovery engine~\cite{Cranmer2023}
        that operates over \rev{random subsets of the data}, mining candidate
        functional forms through evolutionary optimization with no
        prior specification of the mathematical structure.  This
        directly addresses the library design bottleneck that limits
        all fixed-library \sindy{} variants~\cite{Brunton2016,Mangan2017}.

  \item \textbf{A principled symbolic curation pipeline.}
        We introduce a three-stage curation mechanism that
        decomposes raw symbolic expressions into atomic basis
        functions, applies configurable expansion strategies (severe,
        gentle, or hybrid), and prunes multicollinear features \rev{by
        pairwise correlation (or, optionally, variance inflation factor
        (VIF) analysis)} with a simplicity
        bias.  This pipeline transforms unstructured genetic algorithm
        output into a clean, non-redundant linear basis amenable to
        robust sparse regression.

  \item \textbf{\rev{Weak-form sparse identification on the curated library.}}
        \rev{We fit the curated library with the weak form of
        \sindy{}~\cite{MessengerODE2021,Messenger2021}, which integrates the
        equations against test functions instead of differentiating the
        noisy data. A variant that fits the same library with STLSQ
        (\autosindy-STLSQ) shows the effect of this last stage.}

  \item \textbf{Systematic benchmarking against \rev{established} baselines.}
        We evaluate \autosindy{} on \rev{six} canonical nonlinear systems
        spanning oscillatory, limit-cycle, and chaotic regimes \rev{at
        eight noise levels up to 20\% of each variable's standard
        deviation, with 20 runs per system and noise level}, comparing
        against standard \sindy{} with enriched polynomial-Fourier
        libraries, \rev{its weak form,} and standalone symbolic
        regression.  \rev{At every noise level, \autosindy{} passes the
        recovery threshold more often and finds more of the true terms
        than every baseline}, achieving higher predictive accuracy,
        superior generalization to unseen trajectories, and
        \rev{equations close to the ground-truth complexity}.
\end{enumerate}

The remainder of this paper is organized as follows.
Section~\ref{sec:related} reviews related work on sparse
identification, symbolic regression, and hybrid equation discovery
methods.  Section~\ref{sec:method} presents the \autosindy{}
framework in detail.  Section~\ref{sec:experiments} describes the
experimental setup and evaluation metrics.
Section~\ref{sec:results} reports results and comparative analysis.
Section~\ref{sec:discussion} discusses implications, limitations, and
future directions.  Section~\ref{sec:conclusion} concludes.

\section{\label{sec:related}Related Work}

Data-driven equation discovery lies at the intersection of sparse
optimization, evolutionary computation, and scientific machine
learning.  We survey the most relevant lines of work, organized by
methodology and proximity to the \autosindy{} framework.

\subsection{\label{sec:sindy-bg}Sparse Identification of Nonlinear
Dynamics}

The \sindy{} algorithm~\cite{Brunton2016} established the modern
framework for sparse, library-based equation discovery.  Given state
measurements and their time derivatives, \sindy{} assembles a matrix
of candidate nonlinear functions, typically polynomials and
trigonometric terms, and recovers the governing equations as the sparse
solution to an overdetermined regression problem using STLSQ.  The
extension PDE-FIND~\cite{Rudy2017} applied this principle to partial
differential equations, enabling the recovery of spatiotemporal
governing laws such as the Navier-Stokes and Kuramoto-Sivashinsky
equations from data.  The \pysindy{} software
package~\cite{deSilva2020,Kaptanoglu2022} has become \rev{a widely used}
implementation platform, providing modular support for library
construction, optimizer selection, and \rev{numerical differentiation}.

Building on this infrastructure, a rich family of \sindy{} variants has
addressed distinct limitations of the original formulation.  The
\sindy-control extension~\cite{BruntonControl2016} incorporates
external control signals, enabling discovery of actuated dynamics from
input-output data.  \rev{\sindy-PI}~\cite{Kaheman2020} extends the
framework to \rev{implicit dynamics and} rational function nonlinearities through
cross-multiplication.  Autoencoder-\sindy~\cite{Champion2019} jointly
learns a coordinate transformation into a parsimonious latent space
and discovers the sparse governing equations within it.

Noise robustness has been a persistent focus.  The weak-form \sindy{}
approach~\rev{\cite{MessengerODE2021,Messenger2021}} reformulates the regression problem in an
integral form, multiplying both sides by smooth test functions and
integrating over time windows, effectively averaging out noise and
eliminating the need for direct derivative computation.  \esindy{}
(Ensemble-\sindy)~\cite{Fasel2022} takes a complementary statistical
approach, applying bagging to \sindy: an ensemble of models is
identified from subsets of the data, inclusion probabilities are
computed for each candidate library term, and terms with low
inclusion probabilities are thresholded, providing uncertainty
quantification and robustness in the low-data, high-noise
limit.  \rev{Other methods estimate the noise-free states together
with the model, such as modified \sindy{} (mod-\sindy) with automatic
differentiation~\cite{Kaheman2022}, moving-horizon optimization
(DySMHO)~\cite{Lejarza2022} and simultaneous identification and
denoising (SIDDS)~\cite{Hokanson2023}. Sparse data assimilation
(DAHSI)~\cite{Ribera2022} also handles unmeasured variables.}

Model selection has also received sustained attention.  \sindy-AIC~\cite{Mangan2017}
integrates the AIC into the \sindy{} pipeline, ranking candidate models
by their balance of goodness-of-fit and parametric complexity.  The
variant of Dong \emph{et al.}~\cite{Dong2023} further incorporates
group sparsity to identify coefficient-dependent PDEs and introduces a
formal treatment of noise distribution estimation.  SR3~\cite{9194760}
replaces STLSQ with a relaxation that decouples coefficient
shrinkage and thresholding \rev{through an auxiliary variable, and it also
accommodates physical constraints and the trimming of outliers}.  Probabilistic approaches,
including \rev{sparse Bayesian regression with error
bars~\cite{Zhang2018},} sparse Bayesian formulations with spike-and-slab and
horseshoe priors~\rev{\cite{Nayek2021,Hirsh2022}}, \rev{Bayesian uncertainty
quantification for equation learning~\cite{MartinaPerez2021}, Bayesian
model selection~\cite{Niven2024} and a rapid Bayesian method for scarce
and noisy data~\cite{Fung2025},} provide full posterior distributions
over model coefficients\rev{, usually} at the cost of higher computational burden.
\rev{Some of these methods select terms by their probability of
inclusion rather than by thresholding the
coefficients~\cite{Nayek2021,Hirsh2022}.}

\rev{These advances improve how terms are selected and how noise is
handled, but the library itself is still chosen by the user.
ADAM-\sindy~\cite{Viknesh2026}, for example, learns parameters such as
frequencies or exponents inside a family of functions that the user
chooses.}  This library design
problem is the primary motivation for \autosindy. \rev{The library that
\autosindy{} builds can be combined with different last stages, such as
the weak form in \autosindy{} and STLSQ in the variant \autosindy-STLSQ
(Sec.~\ref{subsec:sindy_stage}).}

\subsection{\label{sec:sr-bg}Symbolic Regression}

Symbolic regression (SR) searches the space of all mathematical
expressions representable by a given set of operators and variables,
simultaneously inferring functional form and parameters.  The
foundations of modern SR trace to genetic programming, pioneered by
Koza~\cite{Koza1992}, which represents mathematical expressions as
trees and applies evolutionary operators to iteratively improve a
population of candidate expressions.  The landmark \rev{work of}
Schmidt and Lipson~\cite{Schmidt2009}\rev{, released as the Eureqa software,} established SR as a practical
scientific tool, applying Pareto optimization to jointly minimize
prediction error and expression complexity.

Among current SR tools, \pysr~\cite{Cranmer2023} has established itself
as \rev{a widely used} open-source library for practical symbolic regression in
the sciences.  \pysr{} employs a multi-population evolutionary
algorithm with an evolve-simplify-optimize loop\rev{. P}opulations of
expression trees evolve through mutation and crossover, each
surviving expression is algebraically simplified, and its
numerical constants are refined by nonlinear optimization.  An
adaptive parsimony penalty discourages premature specialization, and a
Julia backend (SymbolicRegression.jl) compiles user-defined operators
into SIMD kernels at runtime.  A recent systematic benchmark by Brum
\emph{et al.}~\cite{Brum2025} evaluated different SR methods on nine canonical
dynamical systems including the Lorenz attractor, Lotka-Volterra
equations, and a suite of epidemiological compartmental models\rev{, two of
them also with added noise}.  \pysr{} was the top-performing \rev{and most robust} method overall,
recovering the correct structural form of all nine systems.

{\color{revcolor}
Other symbolic regression tools take different approaches. AI
Feynman~\cite{Udrescu2020} uses physics-inspired simplifications such
as symmetries and separability. Bayesian symbolic regression, such as
the Bayesian machine scientist~\cite{Guimera2020,Guimera2026}, ranks
expressions by their probability given the data. Exhaustive symbolic
regression (ESR)~\cite{Bartlett2024} enumerates all expressions up to a
given size and ranks them by description length.
Operon~\cite{Burlacu2020} is an efficient genetic-programming library,
and Brush~\cite{Aldeia2026} adds decision-tree-like splits to the
expressions. AI-Descartes~\cite{Cornelio2023} and
AI-Hilbert~\cite{CoryWright2024} combine data with background theory.
}

Despite this success, SR has inherent limitations when applied to
multi-dimensional dynamical systems.  Because SR methods \rev{typically} optimize each
state variable's equation independently, they \rev{struggle to} enforce shared
structure across coupled variables~\cite{Metayer2025}.  The NP-hardness of the general SR
problem~\rev{\cite{virgolin2022symbolicregressionnphard}} means that practical algorithms are heuristic and do not
guarantee convergence to the globally optimal expression.  Recovered
expressions \rev{can be numerically unstable} when integrated as ODEs.  These limitations motivate
the use of SR as a library \emph{generation} tool rather than a
standalone solver.

\subsection{\label{sec:hybrid-bg}Hybrid and Automated Equation
Discovery}

Several prior works have sought to combine the strengths of symbolic
and sparse regression, or to automate the library construction
process.  Fasel \emph{et al.}~\cite{Fasel2022} note that SR
algorithms are inherently imbued with ensembling ideas\rev{, since SR models
are formed by randomly combining building blocks (library terms) and then
recombining those that best model the data. This suggests that} SR and sparse
regression \rev{are} complementary rather than competing approaches.

HAVOK (Hankel Alternative View Of Koopman)~\cite{Brunton2017} \rev{combines}
delay embedding with sparse regression to discover
\rev{forced} linear representations of chaotic systems in \rev{delay} coordinates,
effectively automating a restricted form of basis generation.
Autoencoder-\sindy~\cite{Champion2019} jointly learns nonlinear
coordinate transformations and sparse governing equations, automating
basis discovery for systems with unknown intrinsic coordinates.
\rev{In their review of data-driven digital twins in biomedicine,
M{\'e}tayer \emph{et al.}~\cite{Metayer2025} argue that progress in learning
digital twins will come from hybrid and modular frameworks.}

A critical distinction between these prior works and \autosindy{} is the
treatment of the library as a fully learned, data-driven object with no
domain-knowledge assumptions.  \autosindy{} \rev{uses} symbolic
regression purely as a basis mining tool,
systematically \rev{decomposes and filters} its output through collinearity
analysis, and \rev{hands} the resulting library to \sindy{} for sparse
coefficient recovery, all without any prior knowledge of the system's
mathematical form.

\subsection{\label{sec:deep-bg}Deep Learning and Neural Approaches}

Neural network-based approaches have also been applied to equation
discovery and dynamical system identification.
Physics-Informed Neural Networks (PINNs)~\cite{Raissi2019} embed
known PDEs into the loss function of a neural network, enabling
data-efficient regression for parameterized systems with known
governing structure.  Neural ODEs~\cite{ChenRTQ2018} parameterize
continuous-time dynamics as neural networks, enabling flexible
modeling of black-box dynamical systems but at the cost of
interpretability.  Transformer-based SR models such as
ODEFormer~\cite{dascoli2023odeformersymbolicregressiondynamical} \rev{use} a transformer pretrained on \rev{tens of} millions
of synthetic \rev{examples} to directly predict the symbolic form of a
governing equation from data.
It achieves strong results on low-dimensional benchmark systems\rev{, after
this large-scale pretraining}.
In contrast, \autosindy{} requires no pretraining \rev{and} operates from scratch on
any system for which time-series data can be collected.

\section{\label{sec:method}The \autosindy{} Framework}

\autosindy{} is a three-stage, fully automated pipeline that converts
raw, noisy time-series observations into a sparse, interpretable set
of governing differential equations, without any prior specification
of the system's mathematical form.
Figure~\ref{fig:concept} sketches the high-level concept: noisy
observations feed a hybrid system in which a genetic algorithm
(PySR) and a sparse regressor (SINDy) operate in sequence rather than
in competition.  Figure~\ref{fig:framework_overview} gives a visual overview of the
pipeline.  In Stage~1 (\textsc{Discover}), symbolic regression is
applied to small \rev{random subsets of the training points (chunks)} to mine a diverse pool
of candidate functional forms.  In Stage~2 (\textsc{Curate}), that raw
pool is systematically decomposed, expanded, and filtered by a
collinearity-aware pruning procedure that produces a compact,
well-conditioned library.  In Stage~3 (\textsc{Identify}), \rev{the weak
form of \sindy{} (WSINDy)~\cite{Messenger2021} selects the final terms
from the curated library and fits their coefficients, without
differentiating the noisy data.}
Each stage is described in full below.

\begin{figure*}[t]
	\centering
	\includegraphics[width=\textwidth, height=12cm, keepaspectratio=false]{Framework_revised.pdf}
	\caption{The AutoSINDy hybrid architecture for robust automated discovery of governing equations.    \textbf{\rev{Stage}~1 Discover:} PySR is applied independently to
    multiple \rev{random chunks of the training points}, producing a pool of
    raw symbolic expressions (e.g., $\sin(x)+x^2$, $0.5y\exp(-x)$,
    $\cos(y)$).
    \textbf{\rev{Stage}~2 Curate:} Each expression is symbolically
    decomposed into atomic sub-terms ($\sin(x)+y^2 \to \sin(x),\,y^2$),
    the atoms are sorted by complexity to enforce a simplicity bias,
    and collinear terms (e.g., $\cos(x)y$, which is nearly linearly
    dependent on already-kept terms) are rejected by a VIF/correlation
    redundancy filter.  The result is a compact, stable custom library
    $\boldsymbol{\Theta}(\mathbf{x})$.
    \textbf{\rev{Stage}~3 Identify:} \rev{The weak form of SINDy (WSINDy)
    multiplies each equation by smooth test functions $\varphi_k(t)$ on
    short time windows and integrates over time, so this stage does not
    differentiate the noisy data. Each test function gives one row of
    the linear system $\mathbf{b} = \mathbf{G}\boldsymbol{\xi}$. A sparse
    fit (MSTLS) then drops small terms and refits, giving the
    coefficients $\boldsymbol{\xi}$ of a concise governing equation.}
    \rev{Designed and drawn by the authors in Inkscape. Some
    icons and vector elements were drafted with Claude (Opus 5.5,
    Anthropic) and refined by the authors.}}
	\label{fig:framework_overview}
\end{figure*}

\subsection{Problem Formulation}
\label{subsec:problem_formulation}

We seek to identify the governing ordinary differential equations of
an $n$-dimensional dynamical system
\begin{equation}
  \dot{\mathbf{x}}(t) = \mathbf{f}\!\left(\mathbf{x}(t)\right),
  \qquad \mathbf{x}(0) = \mathbf{x}_0,
  \label{eq:general_system}
\end{equation}
where $\mathbf{x}(t)\in\mathbb{R}^n$ is the state vector and
$\mathbf{f}:\mathbb{R}^n\to\mathbb{R}^n$ is an unknown, potentially
nonlinear vector field.  We observe $m$ snapshots
\begin{equation}
  \mathbf{X} = \bigl[\mathbf{x}(t_1),\ldots,\mathbf{x}(t_m)\bigr]^\top
  \in\mathbb{R}^{m\times n}
\end{equation}
corrupted by additive Gaussian noise of relative amplitude $\sigma$
(measured as a fraction of the per-dimension standard deviation of the
clean trajectory). Time derivatives $\dot{\mathbf{X}}$ are
estimated using PySINDy's \texttt{SmoothedFiniteDifference}
operator~\cite{deSilva2020,Kaptanoglu2022}, which applies a
Savitzky-Golay smoothing filter before finite differencing to
suppress noise amplification in the derivative estimate.

The objective is to recover $\mathbf{f}$ as a sparse, closed-form
symbolic expression.  Following the \sindy{} formulation~\cite{Brunton2016},
this amounts to identifying a sparse coefficient matrix
$\boldsymbol{\Xi}\in\mathbb{R}^{p\times n}$ such that
\begin{equation}
  \dot{\mathbf{X}} \approx \boldsymbol{\Theta}(\mathbf{X})\,\boldsymbol{\Xi},
  \label{eq:sindy_form}
\end{equation}
where $\boldsymbol{\Theta}(\mathbf{X})\in\mathbb{R}^{m\times p}$ is a
library matrix whose $p$ columns are candidate functions evaluated over
all observed states.  Rather than specifying $\boldsymbol{\Theta}$
from prior domain knowledge, \autosindy{} constructs it automatically
from data through the three stages described below.

\subsection{Stage 1: Symbolic Mining via \pysr{} (\textsc{Discover})}
\label{subsec:symbolic_discovery}

The first stage uses \pysr~\cite{Cranmer2023}, a state-of-the-art
multi-population evolutionary symbolic regression tool, to extract a
diverse pool of candidate functional forms directly from trajectory
data.  Two design choices, chunked stochastic sampling and full Pareto
harvesting, ensure that the discovery phase is both computationally
tractable and semantically comprehensive.

\subsubsection{Chunked Stochastic Sampling}

Rather than applying symbolic regression to the full training
trajectory, \autosindy{} operates on $K$ \rev{random subsets (chunks)
of the training points}.  Each chunk is constructed by drawing, without
replacement, an index set $\mathcal{I}_k$ of size $c = \lfloor
m_{\mathrm{train}}/d \rfloor$ from the training samples:
\begin{equation}
  \mathcal{D}_k = \bigl\{(\mathbf{x}_i,\,\dot{\mathbf{x}}_i)
  : i\in\mathcal{I}_k\bigr\}, \quad k=1,\ldots,K,
\end{equation}
where $d$ is the chunk-size divisor and $K$ is the
number of discovery rounds.  This strategy serves
three purposes simultaneously: (i)~each \rev{chunk is small}, making
the evolutionary search substantially more tractable than fitting the
full trajectory, (ii)~different \rev{chunks contain different points},
generating a semantically diverse candidate pool, and (iii)~the random selection
prevents systematic over-representation of any single region of the
attractor, improving the breadth of functional forms discovered.

\subsubsection{Symbolic Regression and Pareto Harvesting}

For each chunk $\mathcal{D}_k$ and each target derivative
$\dot{x}_j$, \autosindy{} invokes \pysr{} with a fixed operator
vocabulary comprising the binary operators and the
unary operators.  Nested
trigonometric compositions (e.g.\ $\sin(\sin(\cdot))$) are explicitly
forbidden via nested constraints, preventing the evolutionary search
from exploiting degenerate identities. These constraints are applied to ensure the search space remains consistent with traditional methods, thereby facilitating a more practical comparison.
 
Crucially, \autosindy{} harvests \emph{every} expression that appears
on the returned Pareto-optimal front, not only the single highest-scoring
model.  Sub-optimal expressions at lower complexity levels frequently
contain structurally correct atomic sub-terms that are globally
valid for the governing equation, even when the full expression is a
poor fit to a particular local chunk.  The union of all Pareto
expressions across all chunks and all target variables defines the raw
symbolic pool:
\begin{equation}
  \mathcal{E}_{\mathrm{raw}}
  = \bigcup_{j=1}^{n}\bigcup_{k=1}^{K}\bigcup_{c}\,e_{j,k,c},
  \label{eq:raw_pool}
\end{equation}
where $e_{j,k,c}$ denotes the Pareto expression of complexity $c$
obtained for target $j$ on chunk $k$.  All expressions are stored as
SymPy~\cite{Meurer2017} symbolic objects for symbolic manipulation in
the subsequent curation stage.

\subsection{Stage 2: Multi-Stage Library Curation (\textsc{Curate})}
\label{subsec:library_curation}

The raw pool $\mathcal{E}_{\mathrm{raw}}$ typically contains hundreds
of compound, overlapping symbolic expressions.  Using them directly as
library columns would produce a severely rank-deficient library matrix
and destabilize the sparse regression.  Stage~2 reduces this pool to a
compact, well-conditioned set of basis functions through three
sequential operations: symbolic decomposition, algebraic expansion,
and collinearity pruning.  The full procedure is given in
Algorithm~\ref{alg:curation}.

\subsubsection{Step 2a: Symbolic Decomposition}

Each expression $e\in\mathcal{E}_{\mathrm{raw}}$ is parsed via SymPy
and decomposed into its additive sub-terms (atoms):
\begin{equation}
  \mathcal{A}_e = \bigl\{a : a \text{ is an additive atom of } e\bigr\}.
\end{equation}
For example, the expression $3x_0^2 + 2\sin(x_1)$ yields atoms
$\{x_0^2,\,\sin(x_1)\}$.  All numeric constant prefactors are
stripped, since \sindy{} learns all coefficients independently during
the regression step.  The full atomic set is formed by
$\mathcal{A} = \bigcup_{e\in\mathcal{E}_{\mathrm{raw}}}\mathcal{A}_e$,
de-duplicated by symbolic equivalence.  This step focuses the library
on unique functional forms rather than rescaled variants of the same
expression.

\subsubsection{Step 2b: Algebraic Expansion}
 
Compound atomic terms may contain grouped polynomial sub-expressions
whose monomials are useful basis functions in their own right.
\autosindy{} supports three configurable expansion strategies applied
to each atom via SymPy's algebraic and trigonometric expansion
routines:
 
\begin{itemize}
  \item \textbf{Severe:} Full polynomial and trigonometric expansion.
        $(x_0+x_1)^2$ becomes $\{x_0^2,\,2x_0 x_1,\,x_1^2\}$ and
        $\sin(x_0+x_1)$ is expanded via the angle-sum identity.  This
        maximizes coverage of monomial basis functions at the cost of
        a larger intermediate pool.
 
  \item \textbf{Gentle:} \rev{Products are multiplied out, but} grouped
        polynomial expressions are preserved intact.
        $(x_0+x_1)^2$ remains as a single atom.  This is the
        default strategy and performs best on systems where the
        true basis functions are grouped.
 
  \item \textbf{Hybrid:} Both the gentle and severe forms of each atom
        are added to the pool, providing maximum coverage at the cost
        of a temporarily larger candidate set before pruning.
\end{itemize}
 
After expansion, all atoms are collected into the augmented candidate
set $\mathcal{A}=\{a_1,\ldots,a_P\}$ and sorted in
\emph{ascending order of SymPy operator count}.  This complexity
ordering enforces a \emph{simplicity bias}: simpler expressions are
always considered before more complex ones in the subsequent pruning
step, ensuring that when two expressions carry equivalent empirical
information, the algebraically simpler one is preferred.

\subsubsection{Step 2c: Collinearity Pruning with Simplicity Bias}

Multicollinearity among library terms is the primary driver of
coefficient instability in sparse regression: near-linearly dependent
columns inflate coefficient variance and frustrate the sparsity
optimizer.  \autosindy{} removes such redundancy through a greedy
forward-selection procedure that processes the complexity-sorted
candidates in order.
 
Let $\mathcal{L}=\emptyset$ be the accepted library, initially empty.
For each candidate atom $a_i\in\mathcal{A}$ (in complexity order):
\begin{enumerate}
  \item Evaluate $\mathbf{v}_i = a_i(\mathbf{X}_{\mathrm{train}})
        \in\mathbb{R}^{m_{\mathrm{train}}}$.
  \item Discard $a_i$ if $\mathrm{std}(\mathbf{v}_i) < \varepsilon$
        (constant or numerically degenerate over the training domain).
  \item Otherwise, compute the pairwise Pearson correlation between
        $\mathbf{v}_i$ and every already-accepted feature
        $\mathbf{v}_j = a_j(\mathbf{X}_{\mathrm{train}})$,
        $a_j\in\mathcal{L}$.  Accept $a_i$ if and only if
        \begin{equation}
          \bigl|\mathrm{corr}\bigl(\mathbf{v}_i,\,\mathbf{v}_j\bigr)\bigr|
          < \rho_{\max}
          \quad \forall\,a_j\in\mathcal{L},
          \label{eq:correlation_criterion}
        \end{equation}
        where $\rho_{\max}$ is the collinearity threshold.
  \item If accepted, append $a_i$ to $\mathcal{L}$.
\end{enumerate}
The complexity-ordered traversal guarantees the simplicity bias: a
complex atom $a_i$ is retained \emph{only} if it provides information
that no simpler already-accepted term can replicate.  This directly
implements Occam's razor at the library level, independently of the
downstream sparse optimizer.
 
An alternative pruning criterion based on the Variance Inflation
Factor (VIF) is also supported.  In this mode, a temporary ordinary
least squares regression is fitted with $a_i$ as the response and all
columns of $\mathcal{L}$ as predictors; the resulting
$R^2_i$ yields
\begin{equation}
  \mathrm{VIF}_i = \frac{1}{1 - R^2_i}.
  \label{eq:vif}
\end{equation}
The atom $a_i$ is accepted if and only if $\mathrm{VIF}_i <
\tau_{\mathrm{VIF}}$.  The VIF criterion detects
\emph{multicollinearity with respect to the entire accepted set}
rather than any single pair, making it more sensitive to jointly
redundant groups of terms.  In our experiments, pairwise correlation
pruning was used as the primary method due to its lower computational
overhead and competitive results; VIF pruning is retained as a
configurable option.
 
The final curated library $\mathcal{L}=\{\ell_1,\ldots,\ell_q\}$,
with $q\le P$, is used to construct the library matrix
$\boldsymbol{\Theta}(\mathbf{X})$.
 
\begin{algorithm}[H]
\caption{Library Curation with Simplicity Bias}
\label{alg:curation}
\SetAlgoLined
\KwIn{Raw pool $\mathcal{E}_{\mathrm{raw}}$, training data
  $\mathbf{X}_{\mathrm{train}}$, threshold $\rho_{\max}$, expansion
  strategy $s\in\{\text{gentle},\text{severe},\text{hybrid}\}$}
\KwOut{Curated library $\mathcal{L}$}
$\mathcal{A}\leftarrow\emptyset$\;
\For{$e\in\mathcal{E}_{\mathrm{raw}}$}{
  Parse $e$ with SymPy; decompose into additive atoms; strip numeric
  \rev{prefactors}\;
  Apply expansion strategy $s$ to each atom\;
  $\mathcal{A}\leftarrow\mathcal{A}\cup\{\text{expanded atoms}\}$\;
}
Sort $\mathcal{A}$ by SymPy operator count (ascending)\;
$\mathcal{L}\leftarrow\emptyset$\;
\For{$a_i\in\mathcal{A}$ (in sorted order)}{
  $\mathbf{v}_i\leftarrow a_i(\mathbf{X}_{\mathrm{train}})$\;
  \lIf{$\mathrm{std}(\mathbf{v}_i)<\varepsilon$}{\textbf{skip}
    (constant or degenerate)}
  \If{$\max_{a_j\in\mathcal{L}}\bigl|\mathrm{corr}
    (\mathbf{v}_i,\mathbf{v}_j)\bigr|<\rho_{\max}$}{
    $\mathcal{L}\leftarrow\mathcal{L}\cup\{a_i\}$\;
  }
}
\Return $\mathcal{L}$
\end{algorithm}

\subsection{Stage 3: Sparse Identification via \texorpdfstring{\rev{Weak-Form}}{Weak-Form}
  \sindy{} (\textsc{Identify})}
\label{subsec:sindy_stage}
 
\subsubsection{Library Matrix Construction}
 
The curated library $\mathcal{L}=\{\ell_1,\ldots,\ell_q\}$ is
augmented with a constant bias term to form a library of $p = q+1$
functions.  The library matrix evaluated over the training data is
\begin{equation}
  \boldsymbol{\Theta}(\mathbf{X}_{\mathrm{train}})
  = \bigl[
    \mathbf{1},\;
    \ell_1(\mathbf{X}_{\mathrm{train}}),\;
    \ldots,\;
    \ell_q(\mathbf{X}_{\mathrm{train}})
  \bigr]
  \in\mathbb{R}^{m_{\mathrm{train}}\times(q+1)},
  \label{eq:feature_matrix}
\end{equation}
where each column is the function evaluated point-wise over all
training observations.

\subsubsection{\texorpdfstring{\rev{Sparse Regression in Weak Form}}{Sparse Regression in Weak Form}}
\label{subsec:weak_form}

{\color{revcolor}
Stage~3 identifies the active terms of the curated library, and their
coefficients, with the weak form of \sindy{} (WSINDy) of Messenger and
Bortz~\cite{MessengerODE2021,Messenger2021}. Instead of fitting
estimated derivatives, it multiplies the equation of each state
variable, $\dot{x}_j = \boldsymbol{\Theta}_j(\mathbf{x})\,\boldsymbol{\xi}_j$,
by smooth test functions $\varphi_k(t)$, each non-zero only on a short
time window, and integrates over time. Integration by parts moves the
time derivative from the data to the test function, so this stage does
not differentiate the noisy data. This gives the linear system
\begin{equation}
  \mathbf{b}_j \approx \mathbf{G}_j\,\boldsymbol{\xi}_j,
  \quad\text{with}\quad
  b_{j,k} = -\int \varphi_k'(t)\,x_j(t)\,dt
  \quad\text{and}\quad
  \mathbf{G}_{j,k} = \int \varphi_k(t)\,
    \boldsymbol{\Theta}_j\bigl(\mathbf{x}(t)\bigr)\,dt,
  \label{eq:weak_form}
\end{equation}
where $\mathbf{G}_{j,k}$ is the $k$-th row of $\mathbf{G}_j$ and the
integrals are evaluated on the training samples. Following Messenger
and Bortz~\cite{Messenger2021}, the test functions are bell-shaped polynomials
whose width and degree are set for each state variable from the corner
point of its Fourier spectrum, where noise starts to dominate, and
their windows are spaced uniformly over the training data. The sparse
coefficients $\boldsymbol{\xi}_j$ are found by their modified
sequentially thresholded least squares (MSTLS), which chooses the
threshold automatically by balancing the fit against the number of
terms. All settings are the defaults of their published code, except
that the library is not rescaled, because their rescaling assumes
polynomial terms.

}

\subsubsection{\texorpdfstring{\rev{The STLSQ Variant (\autosindy-STLSQ)}}{The STLSQ Variant (AutoSINDy-STLSQ)}}

\rev{In the variant \autosindy-STLSQ, Stages~1 and~2 are unchanged, and}
the sparse coefficient matrix $\hat{\boldsymbol{\Xi}}$ is \rev{instead
obtained by fitting the estimated derivatives, solving} the
$\ell_0$-penalized regression problem
\begin{equation}
  \hat{\boldsymbol{\Xi}}
  = \argmin_{\boldsymbol{\Xi}}
    \bigl\|\dot{\mathbf{X}}_{\mathrm{train}}
    - \boldsymbol{\Theta}(\mathbf{X}_{\mathrm{train}})\boldsymbol{\Xi}
    \bigr\|_F^2
    + \lambda\|\boldsymbol{\Xi}\|_0,
  \label{eq:sparse_optimization}
\end{equation}
where $\|\cdot\|_F$ is the Frobenius norm.
 
Two sparse optimizers are supported.  \textbf{Sequential Thresholded
Least Squares (STLSQ)}~\cite{Brunton2016} is the default\rev{. It}
alternates between a least-squares fit\rev{, with a small ridge penalty as in \pysindy{},} and a hard coefficient threshold at
level $\lambda$ until the active set converges.  The threshold
$\lambda$ acts as the principal sparsity-promoting hyperparameter.  Alternatively, \textbf{Sparse Relaxed
Regularized Regression (SR3)}~\cite{9194760} reformulates the problem
with a relaxation that introduces an auxiliary variable,
decoupling the fitting and thresholding sub-problems.  \rev{SR3 is
available as a configurable option, with relaxation parameter $\nu$ and a
fixed maximum number of iterations.}

\subsubsection{Unified vs.\ Per-Variable Library Strategies}
 
\autosindy{} supports two strategies for library construction and
fitting, offering a trade-off between cross-equation consistency and
per-equation specificity.
 
\paragraph{Separate libraries (default).}
For each target derivative $\dot{x}_j$, Stage~1 runs PySR
independently and Stage~2 curates a variable-specific library
$\mathcal{L}_j$.  A separate \sindy{} model is then fitted
for each $j$, solving
$\dot{x}_j \approx \boldsymbol{\Theta}_j(\mathbf{X})\,\boldsymbol{\xi}_j$.
This strategy allows each equation to contain only the functional
forms that are locally relevant, and typically produces the most
parsimonious individual equations.
 
\paragraph{Unified library.}
Alternatively, Stage~1 harvests expressions for all target variables
simultaneously; Stage~2 pools and curates them into a single shared
library $\mathcal{L}$; and a single \sindy{} model is fitted across
all state variables jointly via Eq.~\eqref{eq:sindy_form}.  The
unified strategy is better suited to tightly coupled systems where the
same functional forms appear in multiple equations, as it enforces
structural consistency at the cost of including terms that may be
irrelevant for some equations.
 
In our experiments, the separate-library strategy is used as the
default because it produces consistently lower complexity and more
accurate identification across the benchmark suite.

\subsection{Complete Pipeline Summary}
\label{subsec:pipeline_summary}
 
The full \autosindy{} procedure is summarized in
Algorithm~\ref{alg:autosindy}.  %
 
\begin{algorithm}[H]
\caption{\autosindy{}: Automated Sparse System Identification}
\label{alg:autosindy}
\SetAlgoLined
\KwIn{Noisy observations $\mathbf{X}$, estimated derivatives
  $\dot{\mathbf{X}}$, hyperparameters $K,d,\rho_{\max}$}
\KwOut{Sparse symbolic model $\hat{\boldsymbol{\Xi}}$ and governing
  equations $\hat{\mathbf{f}}$}
\tcp{Stage 1: Symbolic Discovery}
$\mathcal{E}_{\mathrm{raw}}\leftarrow\emptyset$\;
\For{$k=1,\ldots,K$}{
  Sample random chunk $\mathcal{D}_k$ of size
  $\lfloor m_{\mathrm{train}}/d\rfloor$\;
  \For{$j=1,\ldots,n$}{
    Run \pysr{} on $(\mathcal{D}_k,\,\dot{x}_j)$; harvest full Pareto
    front $\mathcal{P}_{jk}$\;
    $\mathcal{E}_{\mathrm{raw}}\leftarrow\mathcal{E}_{\mathrm{raw}}
    \cup\mathcal{P}_{jk}$\;
  }
}
\tcp{Stage 2: Library Curation (Algorithm~\ref{alg:curation})}
$\mathcal{L}\leftarrow\textsc{Curate}
(\mathcal{E}_{\mathrm{raw}},\,\mathbf{X}_{\mathrm{train}},\,
\rho_{\max})$\;
Build library matrix
$\boldsymbol{\Theta}(\mathbf{X}_{\mathrm{train}})$
from $\mathcal{L}\cup\{1\}$\;
\tcp{Stage 3: \rev{Weak-Form} Sparse Identification}
\For{\rev{$j=1,\ldots,n$}}{
  \rev{Build $\mathbf{b}_j$ and $\mathbf{G}_j$ via Eq.~\eqref{eq:weak_form}}\;
  \rev{Solve $\mathbf{b}_j\approx\mathbf{G}_j\boldsymbol{\xi}_j$ with
  MSTLS $\to\hat{\boldsymbol{\xi}}_j$}\;
}
\Return $\hat{\boldsymbol{\Xi}}$,\;
\hspace{2.5em}$\hat{\mathbf{f}}(\mathbf{x}) =
\boldsymbol{\Theta}(\mathbf{x})\hat{\boldsymbol{\Xi}}$
\end{algorithm}

\section{\label{sec:experiments}Experimental Setup}

Evaluating a symbolic discovery framework requires more than measuring
predictive accuracy on training data: a method that memorizes the
training trajectory adds no scientific value, and a method that
produces interpretable equations which immediately diverge under
integration is not physically meaningful.  Our experimental design
therefore makes two deliberate choices.  First, all generalization
metrics are computed on a \emph{completely unseen} validation
trajectory generated from a different initial condition with zero
noise, ensuring that reported scores reflect the ability to capture
true governing physics rather than noise-specific patterns.  Second,
we evaluate both \emph{derivative prediction} (a local, instantaneous
test of structural correctness) and \emph{simulation} (a
global, integral test of dynamical stability), since the two can
diverge dramatically: an equation may fit local derivatives well yet
diverge catastrophically under forward integration.   The full
experimental protocol is described below.

\subsection{Benchmark Dynamical Systems}
\label{subsec:systems}
 
We evaluate \autosindy{} on six canonical nonlinear dynamical systems
spanning a wide range of structural complexity and dynamic character
(Table~\ref{tab:benchmark_systems}).  The true canonical complexity, 
measured as the total symbolic operator count of the fully expanded
governing equations, ranges from 3 for the simple harmonic oscillator
to 15 for the Complex Lorenz system, which couples chaotic dynamics
with a transcendental cross-term $\gamma x_1\sin(x_0+x_2)$.

\begin{table*}[htbp]
\centering
\caption{Benchmark dynamical systems, governing equations, key
  dynamical characteristics, and ground-truth canonical complexity
  (SymPy operator count of the expanded form).}
\label{tab:benchmark_systems}
\renewcommand{\arraystretch}{1.4} %
\setlength{\tabcolsep}{12pt} %
\resizebox{\textwidth}{!}{%
\begin{tabular}{@{}lclcc@{}}
\toprule
\textbf{System} & $n$ & \textbf{Governing Equations}
  & \textbf{Characteristics} & \textbf{True Complexity} \\
\midrule
Harmonic Oscillator
  & 2
  & $\begin{aligned} \dot{x}_0 &= x_1 \\[-8pt] \dot{x}_1 & = -k_1 x_0 - k_2 x_1 \end{aligned}$
  & \rev{Damped oscillation}
  & 3 \\
Damped Pendulum
  & 2
  & $\begin{aligned} \dot{x}_0 &= x_1 \\[-8pt] \dot{x}_1 &= -bx_1-c\sin(x_0) \end{aligned}$
  & \rev{Damped oscillation}
  & 4 \\
Modulated Oscillator
  & 2
  & $\begin{aligned} \dot{x}_0 &= x_1 \\[-8pt] \dot{x}_1 &= -bx_1\cos(x_0)-kx_0 \end{aligned}$
  & \rev{Damped oscillation}
  & 5 \\
Van der Pol
  & 2
  & $\begin{aligned} \dot{x}_0 &= x_1 \\[-8pt] \dot{x}_1 &= \mu(1-x_0^2)x_1-x_0 \end{aligned}$
  & \rev{Limit cycle}
  & 6 \\
Duffing Oscillator
  & 2
  & $\begin{aligned} \dot{x}_0 &= x_1 \\[-8pt] \dot{x}_1 &= -\delta x_1-\alpha x_0-\beta x_0^3 \end{aligned}$
  & \rev{Decays to one of two equilibria}
  & 6 \\
Complex Lorenz
  & 3
  & $\begin{aligned} \dot{x}_0 &= \sigma(x_1-x_0) \\[-8pt] \dot{x}_1 &= x_0(\rho-x_2)-x_1 \\[-8pt] \dot{x}_2 &= x_0x_1-\beta x_2+\gamma x_1\sin(x_0+x_2) \end{aligned}$
  & Chaotic
  & 15 \\
\bottomrule
\end{tabular}%
}
\end{table*}

\subsection{Data Generation and Noise Protocol}
\label{subsec:data}
 
\begin{figure}[t]
  \centering
  \includegraphics[width=\columnwidth]{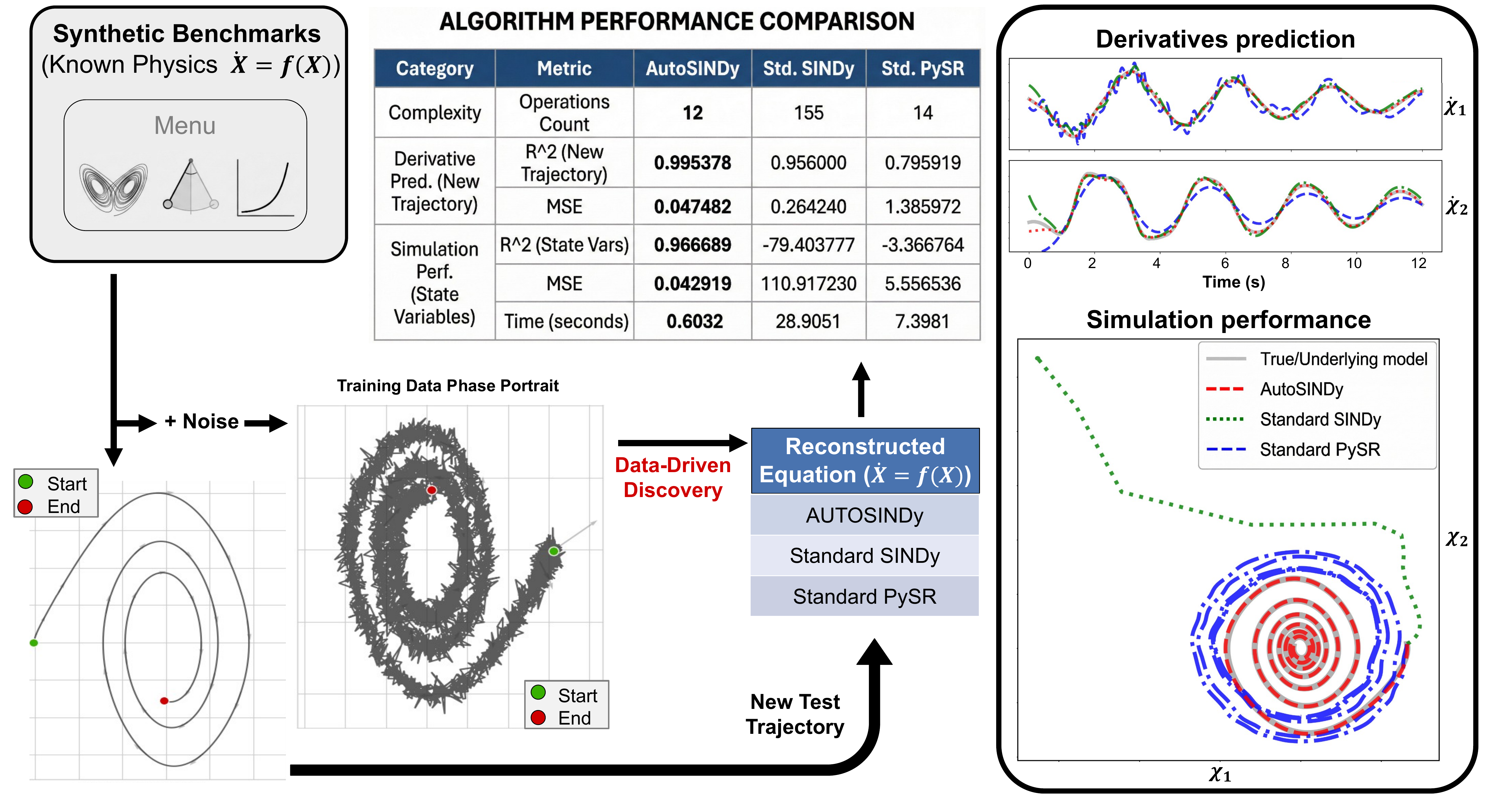}
  \caption{Experimental protocol and an illustrative example.  A trajectory is generated from a system with known physics and then
    corrupted by Gaussian noise to produce the training set (centre
    left).  \rev{Each of the five methods (Sec.~\ref{subsec:baselines})}
    independently \rev{attempts} to recover the governing equations
    from this noisy data.  Discovered models are validated on a
    \emph{clean} trajectory from a previously unseen initial
    condition: once for derivative prediction accuracy and once for
    simulation.  \rev{The figure shows three of the
    methods.}  The right panel shows a representative
    single-trial comparison on the harmonic oscillator, where
    \autosindy{}\rev{-STLSQ (labelled \autosindy{} in the figure)}
    achieves the best accuracy at a fraction of the
    complexity of Standard \sindy.}
  \label{fig:experimental_setup}
\end{figure}

Figure~\ref{fig:experimental_setup} illustrates the complete data
generation and evaluation pipeline.  For each system, a training
trajectory of 5000 uniformly spaced snapshots is simulated via
SciPy's \texttt{solve\_ivp} \rev{(RK45 with the default tolerances,
relative $10^{-3}$ and absolute $10^{-6}$)}.  Additive Gaussian noise
is injected relative to each state variable's standard deviation:
\begin{equation}
  \mathbf{X}_{\mathrm{noisy}} = \mathbf{X}_{\mathrm{clean}}
  + \boldsymbol{\varepsilon},\quad
  \varepsilon_{ij}\sim\mathcal{N}\!\bigl(0,\,
  \sigma^2\mathrm{Var}(X_{\cdot j})\bigr),
\end{equation}
at \rev{eight} levels $\sigma\in\{0,0.01,0.02,0.03,0.04,0.05,\rev{0.10,0.20}\}$ representing the
noise-to-signal ratio.  \rev{The levels are spaced finely up to 0.05, a range
that is already demanding because estimating the derivatives amplifies the
noise (see below). Two higher levels, 0.10 and 0.20, test much stronger
noise.}  Each
level is repeated across \rev{20} independent random seeds
($s\in\{32,\ldots,\rev{51}\}$), giving
$6\text{ systems}\times \rev{8}\text{ noise levels}\times
\rev{20}\text{ seeds}=\rev{960}$ trials per method, and \rev{4800} total runs.
\rev{The training trajectory starts from the same initial condition in every
trial, so the trials differ only in the noise and in the random choices of
the methods.}

 The qualitative impact of noise on raw signal fidelity is illustrated
in Appendix~\ref{app:noise_levels}, which shows the state trajectories,
numerically estimated derivatives, and phase portraits for the damped
pendulum at noise levels $\sigma \in \{0.00, 0.01, 0.05\}$. 
At $\sigma = 0.05$, the raw derivative signal is heavily corrupted, 
motivating the need for robust library selection and regularization 
as implemented in AutoSINDy.

{\color{revcolor}
The noise level $\sigma$ thus refers to the states. At $\sigma = 0.05$, the noise
on each state variable has a standard deviation of 5\% of that variable's standard
deviation over the whole trajectory. All methods except Weak \sindy{}, however,
fit derivatives estimated from the noisy states (\autosindy{} only in Stage~1),
and estimating the derivatives amplifies the noise. For the harmonic oscillator, 1\% noise on the states gives an error in the
estimated derivatives of 45\% of the derivative signal, and 5\% noise gives 224\%.
Appendix~\ref{app:noise_levels} (Table~\ref{tab:derivative_noise}) gives these
values for all six systems.

}
 
The \emph{training trajectory}
is noisy and split 80/20 using a \emph{middle split}: the central 80\%
of the time series forms the training set, while the flanking 10\%
segments form the hold-out test, demanding accuracy in both
interpolation and boundary extrapolation.  

A second, \emph{fully independent validation trajectory} is generated
from the same system with distinct initial conditions and
\emph{zero} measurement noise ($\sigma=0$). This clean trajectory is
never exposed to the discovery or fitting stages; it is used
exclusively to assess whether the discovered model captures the true
underlying physics rather than characteristics of the training noise.
Simulation performance is assessed by numerically integrating the
discovered model forward in time from the validation trajectory's
initial condition and comparing against the corresponding clean
ground-truth trajectory.

\subsection{Baseline Methods}
\label{subsec:baselines}

\textsc{AutoSINDy} is benchmarked against \rev{three} established methods\rev{,
and against its variant \autosindy-STLSQ}:

\begin{enumerate}
  \item \textbf{Standard SINDy with enriched library}: Uses a fixed
    candidate library combining polynomial terms up to degree~3 and
    Fourier basis functions with one frequency:
    $\mathcal{F}_{\mathrm{std}} = \mathrm{PolyLib}(d{=}3)
    \oplus [\mathrm{PolyLib}(d{=}3) \times \mathrm{FourierLib}(f{=}1)]$.
    This large, generic library \rev{contains polynomials, sines and
    cosines and their products, close to the terms that \autosindy's
    operators can build, and it does not assume the form of any
    particular system}. The \rev{same}
STLSQ optimizer \rev{as in \autosindy-STLSQ} is used to ensure
that performance differences reflect library construction rather than
the optimizer. \rev{Without noise, this baseline reaches a new-trajectory
$R^2$ above 0.99 on five of the six systems. It fails on the Duffing
oscillator, whose training data cover a narrow range on which many
library terms are almost combinations of others
(Appendix~\ref{app:sindy_check}).}

  \item \rev{\textbf{Weak SINDy}: the library $\mathcal{F}_{\mathrm{std}}$
    of Standard \sindy{} with the weak-form last stage of \autosindy{}, so
    it does not differentiate the noisy data. It differs from \autosindy{}
    only in the symbolic mining and library construction. Without noise, it
    reaches a new-trajectory $R^2$ above 0.99 on three of the six systems and
    fails on the harmonic, modulated and Duffing oscillators
    (Appendix~\ref{app:sindy_check}).}

  \item \textbf{Standard PySR }: Direct symbolic regression
    applied to the full training set, using the same operator vocabulary
    and hyperparameters as the \textsc{AutoSINDy} discovery stage, but
    without the curation or sparse-identification stages. The best
    equation on the Pareto front is selected independently for each
    state derivative\rev{, by \pysr's default rule, which balances error
    and expression size}.

  \item \rev{\textbf{\autosindy-STLSQ}: \autosindy{} with the STLSQ last
    stage on the estimated derivatives. In each run it uses the same
    library as \autosindy{}, so comparing the two shows the effect of the
    last stage.}
\end{enumerate}

\subsection{Evaluation Metrics}
\label{subsec:metrics}
 
Model performance is assessed with \rev{six} complementary criteria that
together probe \emph{local structural correctness} (derivative
prediction), \emph{global dynamical validity} (\rev{simulation}),
\emph{equation recovery} (\rev{the share of runs that reach $R^2 > 0.99$
on the new trajectory}), \rev{\emph{term-level recovery} (the true
positivity ratio),} \emph{symbolic parsimony}
(operator-level complexity), and \emph{computational efficiency}
(wall-clock timing).  Using all \rev{six} is essential because the criteria
can diverge dramatically: a model may achieve high derivative $R^2$
yet diverge catastrophically under integration, or may produce a
compact equation that is nonetheless structurally incorrect.  No
single score is sufficient on its own. \rev{The $R^2$ on the new
trajectory shows how well a model predicts dynamics it has not seen, the
complexity shows how short the model is, and the true positivity ratio
shows whether its terms are the true ones.}

\subsubsection{Derivative Prediction Accuracy}
 
The coefficient of determination $R^2$ and mean squared error (MSE)
between the \rev{reference} derivatives and the model's instantaneous
predictions are computed on both the held-out test segment of the
training trajectory and the independent clean validation trajectory.
\rev{On the validation trajectory, the reference derivatives are estimated
from the noise-free states in the same way as the training derivatives
(Savitzky--Golay filter, then finite differences). $R^2$ is the average of
the $R^2$ values of the $n$ state variables, and the MSE is taken over all
$N$ time points and $n$ state variables:}
\begin{equation}
  \textcolor{revcolor}{R^2 = \frac{1}{n}\sum_{j=1}^{n}\left(1 -
  \frac{\bigl\|\dot{\mathbf{x}}_j - \hat{\dot{\mathbf{x}}}_j\bigr\|^2}
       {\bigl\|\dot{\mathbf{x}}_j - \bar{\dot{x}}_j\bigr\|^2}\right)},
  \qquad
  \mathrm{MSE} =
  \frac{1}{Nn}
  \bigl\|\dot{\mathbf{X}} - \hat{\dot{\mathbf{X}}}\bigr\|_F^2.
  \label{eq:derivative_metrics}
\end{equation}
Scores on the validation trajectory are the primary indicator of
\rev{how well a model generalizes. The model must reproduce the
derivatives on a noise-free trajectory from a different initial
condition, which it has not seen.}
Scores on the test segment of the training trajectory serve as a
secondary check for overfitting.
\rev{Table~\ref{tab:system_master_summary} gives each $R^2$, of the derivatives
and of the simulations, as the normalized root-mean-square error
$\mathrm{NRMSE} = \sqrt{1 - R^2}$. For a single state variable, this is the
root-mean-square error divided by the standard deviation of the reference.
NRMSE is 0 for a perfect model, and $R^2 = 0.99$ corresponds to an NRMSE of 0.1.}

\subsubsection{Equation Recovery Rate}
 
For each system and noise level, we report the fraction of trials
(across seeds) in which the discovered model achieves $R^2 > 0.99$
on the clean validation trajectory.  This strict binary threshold
\rev{tests whether a model reproduces the dynamics of a trajectory it has
not seen.}  The recovery rate
therefore measures the fraction of trials in which the discovery
pipeline succeeds in \rev{finding a model that generalizes}, rather than merely
fitting the training data.

{\color{revcolor}
\subsubsection{Term-Level Recovery}
\label{subsec:tpr}

The recovery rate does not check the individual terms of a model. For
the benchmark systems, whose true equations are known, we therefore also
report the true positivity ratio of Messenger and
Bortz~\cite{Messenger2021},
\begin{equation}
  \mathrm{TPR} = \frac{n_{\mathrm{found}}}
  {n_{\mathrm{found}} + n_{\mathrm{missed}} + n_{\mathrm{extra}}},
  \label{eq:tpr}
\end{equation}
counted over all equations of a model. Here $n_{\mathrm{found}}$ is the
number of true terms that the model contains, $n_{\mathrm{missed}}$ the
number of true terms it lacks, and $n_{\mathrm{extra}}$ the number of its
terms that are not in the true equations. The TPR equals 1 only when a
model contains exactly the true terms, and it decreases with every
missed or extra term. Found and true terms are compared by their values
on the noise-free training states, so that a term written in a different
form still counts as the true term.
}

\subsubsection{Simulation Stability}

The discovered governing equations are numerically integrated forward
from the validation trajectory's initial condition \rev{over 2 seconds of
simulated time (200 samples)} using SciPy's
stiff Radau ODE solver.  Simulated trajectories are compared against
the corresponding noise-free ground-truth trajectory, with $R^2$ and
MSE computed over the state variables.  Integration is terminated
early if any state variable exceeds $100\times$ the ground-truth
amplitude, which serves as a divergence detection criterion\rev{, or if it
takes longer than 200\,s of computing time}.
\rev{Trials in which the integration crashes (NaN or Inf values, or early
termination by either criterion) are recorded as hard failures and counted in
Fig.~\ref{fig:sim_failures}. Figure~\ref{fig:tiers} counts them as failed, and
Fig.~\ref{fig:noise_robustness} shows them in its failed band.
Table~\ref{tab:system_master_summary} gives them a simulation $R^2$ of $-1$
(an NRMSE of 1.414), and the simulation rows of Fig.~\ref{fig:dashboard}
leave them out. Many runs that did not crash score below $-1$, so both
choices make the methods look better than they are.}  Simulation $R^2$ therefore captures whether the
discovered equation is not only locally accurate in its derivative
predictions but also globally stable and physically coherent when
used as a predictive model.

\subsubsection{Symbolic Parsimony}
 
We measure equation complexity as the total operator count of the
fully algebraically expanded governing equations, computed via
SymPy's \texttt{count\_ops} after applying \texttt{sympy.expand()}.
Formally, for a discovered equation $\hat{f}$ with canonical
(expanded) form $\tilde{f} = \mathrm{expand}(\hat{f})$, we define
\begin{equation}
  C(\hat{f}) = \texttt{count\_ops}\!\bigl(\tilde{f}\bigr),
  \label{eq:complexity}
\end{equation}
where \texttt{count\_ops} tallies every arithmetic and transcendental
operation in the expression tree (additions, multiplications,
exponentiations, $\sin$, $\cos$, etc.).
 
\paragraph{Why not the standard SINDy term count?}
The conventional complexity metric in sparse identification is the
$\ell_0$ norm of the coefficient vector which is the number of nonzero
library terms.  This measure assigns equal complexity to $x_1$ and
$x_1^2 x_2 x_3$, treating a simple linear variable and a fourth-degree
monomial as indistinguishable in cost.  Such a metric cannot
differentiate a genuinely parsimonious physical model from a bloated
one that happens to use few, but individually complex, terms.
Operator count, by contrast, grows proportionally with the actual
arithmetic operations required to evaluate the expression, providing a
complexity measure that is directly consistent with the computational,
analytic, and interpretability cost of the discovered equation.
 
\paragraph{Why expand before counting?}
\pysr{} naturally returns factored expressions, e.g.\
$\mu(1-x_0^2)x_1$ for the Van der Pol nonlinearity, while \sindy{}
always returns expanded polynomial forms, e.g.\
$\mu x_1 - \mu x_0^2 x_1$.  Both representations encode exactly the
same physics, yet their raw operator counts differ.  Measuring
complexity on the as-written form would therefore systematically and
unfairly favour \pysr{} over \sindy{}-based methods with no
scientific justification.  Applying \texttt{sympy.expand()} before
counting places all methods on an identical canonical representation,
and the resulting operator count can be directly compared against the
known ground-truth complexity of the benchmark systems reported in
Table~\ref{tab:benchmark_systems}.
\rev{The \sindy-based methods write a coefficient for every term, also where
the true coefficient is one, so a model with exactly the true terms counts
one or two operators more than the ground truth (8 against 6 for the
Van~der~Pol oscillator).}
 
\paragraph{Sign normalization.}
An expression whose leading term is negative is negated before
counting, using SymPy's \texttt{could\_extract\_minus\_sign},
since $-2x_0 - 3x_1$ and $2x_0 + 3x_1$ represent equally complex
models under a sign flip of the coefficient.  This prevents the
canonical form from artificially inflating complexity for equations
whose dominant term happens to be negative.

\subsubsection{Computational Efficiency}
 
We report wall-clock timing separately for two stages to allow
independent interpretation:
 
\begin{enumerate}
  \item Discovery time: the total elapsed time for
    Stage~1 (\pysr{} across all chunks and state variables),
    Stage~2 (curation), and Stage~3 (\rev{the final sparse fit}).
    This is the time a practitioner must wait before a governing
    equation is available.
 
  \item Simulation time: the time required to numerically
    integrate the discovered governing equations forward over the
    simulation horizon using the Radau solver.  This reflects the
    practical cost of using the equation for downstream prediction or
    control, and is directly affected by equation complexity: bloated
    equations with many terms produce stiffer ODE systems that require
    finer time-stepping, leading to substantially longer integration
    times even when two equations nominally achieve similar $R^2$
    scores.
\end{enumerate}
 
These two timing quantities capture complementary aspects of
computational cost and are not aggregated, since a fast discovery with
a complex equation and a slow discovery with a parsimonious equation
represent qualitatively different trade-offs. \rev{All times were measured
on one laptop, with three runs in parallel.}

\subsection{Implementation Details}
\label{subsec:hyperparams}

\textsc{AutoSINDy} is implemented in Python using \textsc{PySINDy}
v\rev{2.1.0}~\cite{Kaptanoglu2022,deSilva2020}, \textsc{PySR}
v\rev{1.5.10}~\cite{Cranmer2023}, and \textsc{SymPy} v1.14.0~\cite{Meurer2017}
for symbolic manipulation. 
PySR is run in
deterministic serial mode (\texttt{parallelism="serial"},
\texttt{deterministic=True}) to ensure reproducibility. Source code
and all experimental configurations are available at
\url{https://github.com/mabasiri95/AutoSINDy}.
\rev{The authors used Claude (Opus 5.5, Anthropic) to assist with coding
and writing, and they reviewed and verified all of its output.}

{\color{revcolor}
All experiments use the same settings for every system and noise level,
and no setting was tuned per system or per noise level. Stage~1 runs
\pysr{} with the operators $+$, $-$, $\times$, $\sin$, $\cos$, square and
cube (no $\sin$ or $\cos$ inside $\sin$ or $\cos$), 30 iterations,
16 populations and a maximum expression size of 8, the same as Standard
\pysr{}, on $K = 3$ chunks of 400 random training points ($d = 10$).
Appendix~\ref{app:settings} tests other chunks. Stage~3 uses the defaults of Messenger and Bortz's
code (without library rescaling), as does Weak \sindy{}.
\autosindy-STLSQ uses STLSQ with $\lambda = 0.21$ and ridge
$\alpha = 0.05$, as does Standard \sindy{}. Derivatives, where needed,
come from \pysindy's default, a Savitzky--Golay filter (11 points,
cubic) followed by finite differences.

We checked the maximum expression size of \pysr{} afterwards, with sizes
12 and 16, on the six systems at $\sigma = 0$ and 0.05 with seeds 32--36
(30 runs per size and noise level). Without noise, the mean TPR of
\autosindy{} (Eq.~\eqref{eq:tpr}) was 0.88, 0.89 and 0.90 with sizes 8,
12 and 16. At $\sigma = 0.05$ it was 0.81, 0.70 and 0.70, and the lowest
new-trajectory $R^2$ was 0.90, 0.95 and 0.36. Larger expressions thus
help a little without noise but add wrong terms with noise.
}

\section{\label{sec:results}Results}

We organize the results around four core claims: (i)~\textsc{AutoSINDy}
achieves superior derivative prediction accuracy across systems and noise
levels; (ii)~it dramatically outperforms baselines in
simulation stability; (iii)~it discovers equations of near-ground-truth
complexity while \rev{the} baselines systematically over- or under-compress;
and (iv)~these gains are achieved at a modest computational overhead.
Figure~\ref{fig:experimental_setup} provides an illustrative example that motivates the full quantitative analysis and answering the following questions:  Does \autosindy{}
recover the correct governing structure? Do discovered models
remain stable \rev{in} simulation? How parsimonious
are the recovered equations?  How does noise robustness compare
across methods?

\subsection{Overall Reliability: Derivative and Simulation Accuracy}
\label{subsec:results_tiers}

\begin{figure}[t]
  \centering
  \includegraphics[width=\columnwidth]{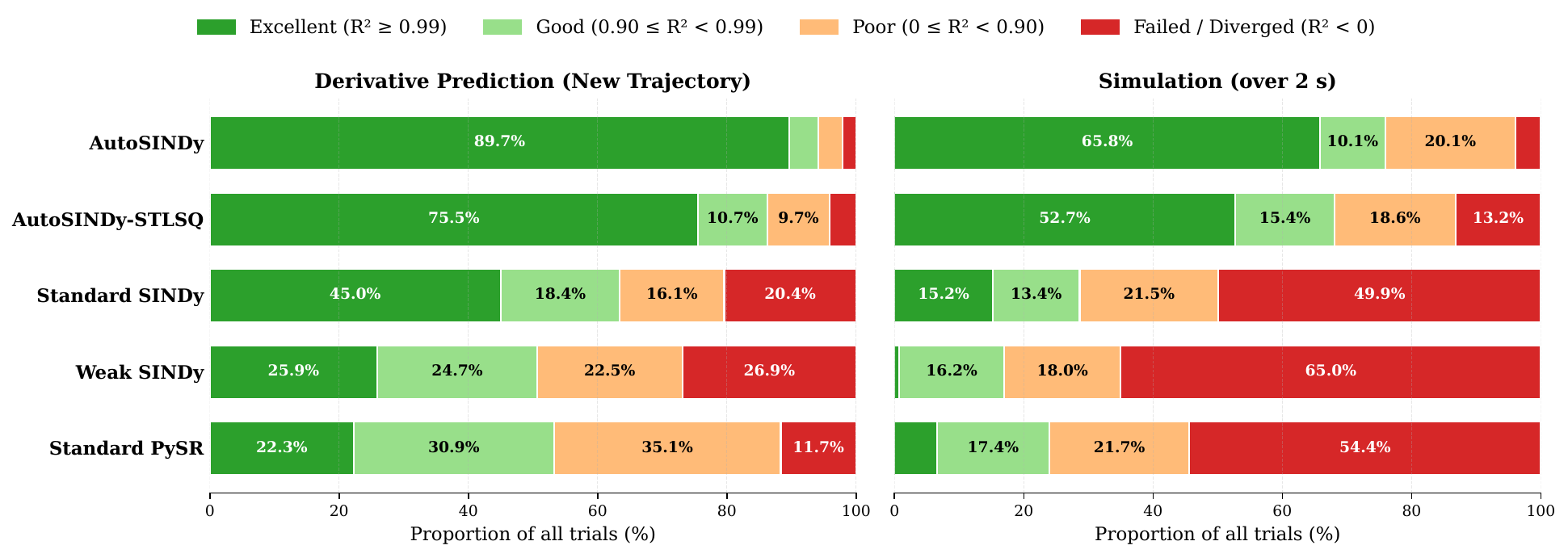}
  \caption{%
    Overall reliability of the \rev{five} methods, aggregated over all six
    dynamical systems and all \rev{eight} noise levels ($N=\rev{960}$ trials per
    method). Horizontal bars show the proportion of trials in each
    accuracy tier for derivative prediction \rev{(left)} and
    simulation \rev{(right)}. \textsc{AutoSINDy}
    achieves the highest proportion of excellent outcomes in both
    categories while maintaining the lowest failure rate.%
  }
  \label{fig:tiers}
\end{figure}

Figure~\ref{fig:tiers} provides the clearest macroscopic
summary of relative performance.  Trials are classified into four
tiers based on \rev{their $R^2$ on the new trajectory (derivative
prediction) and in simulation}: Excellent ($R^2\geq0.99$),
Good ($0.90\leq R^2<0.99$), Poor ($0\leq R^2<0.90$), and
Failed ($R^2<0$).

\paragraph{Derivative prediction.}
\textsc{AutoSINDy} places \textbf{\rev{89.7\%}} of all trials in the
Excellent tier for derivative prediction on the clean validation
trajectory, compared to \rev{75.5\% for \autosindy-STLSQ, 45.0\%} for Standard SINDy, \rev{25.9\% for Weak SINDy} and only \rev{22.3\%} for
Standard PySR. Standard SINDy, despite its large fixed library, places
\rev{20.4\%} of trials in the Failed tier\rev{,} which is a consequence of the
ill-conditioning induced by the high-dimensional enriched library at
elevated noise levels. Standard PySR is unable to achieve excellent
accuracy in the majority of cases (\rev{77.7\%} of trials fall below $R^2 =
0.99$), reflecting its susceptibility to noise-driven over-complexity.

\paragraph{Simulation stability.}
The gap between methods widens dramatically when evaluating
simulation \rev{accuracy} (Fig.~\ref{fig:tiers}, \rev{right panel}).
\textsc{AutoSINDy} achieves Excellent simulation performance in
\textbf{\rev{65.8\%}} of all trials \rev{and \autosindy-STLSQ in 52.7\%}, while Standard SINDy, \rev{Standard PySR and Weak SINDy}
achieve only \rev{15.2\%, 6.6\% and 0.7\%}. Most strikingly, Standard
SINDy \emph{fails} (\rev{$R^2 < 0$}) in \rev{49.9\%} of trials\rev{,} Standard PySR in
\rev{54.4\% and Weak SINDy in 65.0\%}, underscoring that good derivative prediction alone does not
guarantee stable forward integration. \textsc{AutoSINDy} fails in only
\rev{4.0\% of trials (38 of 960)}, and, as detailed next, these failures are concentrated
\rev{on the Duffing oscillator (18) and} at the highest noise levels \rev{(26 at $\sigma = 0.10$ and $0.20$)}.

\subsection{Noise Robustness}
 \label{subsec:results_noise}

\begin{figure*}[t]
  \centering
  \includegraphics[width=\textwidth]{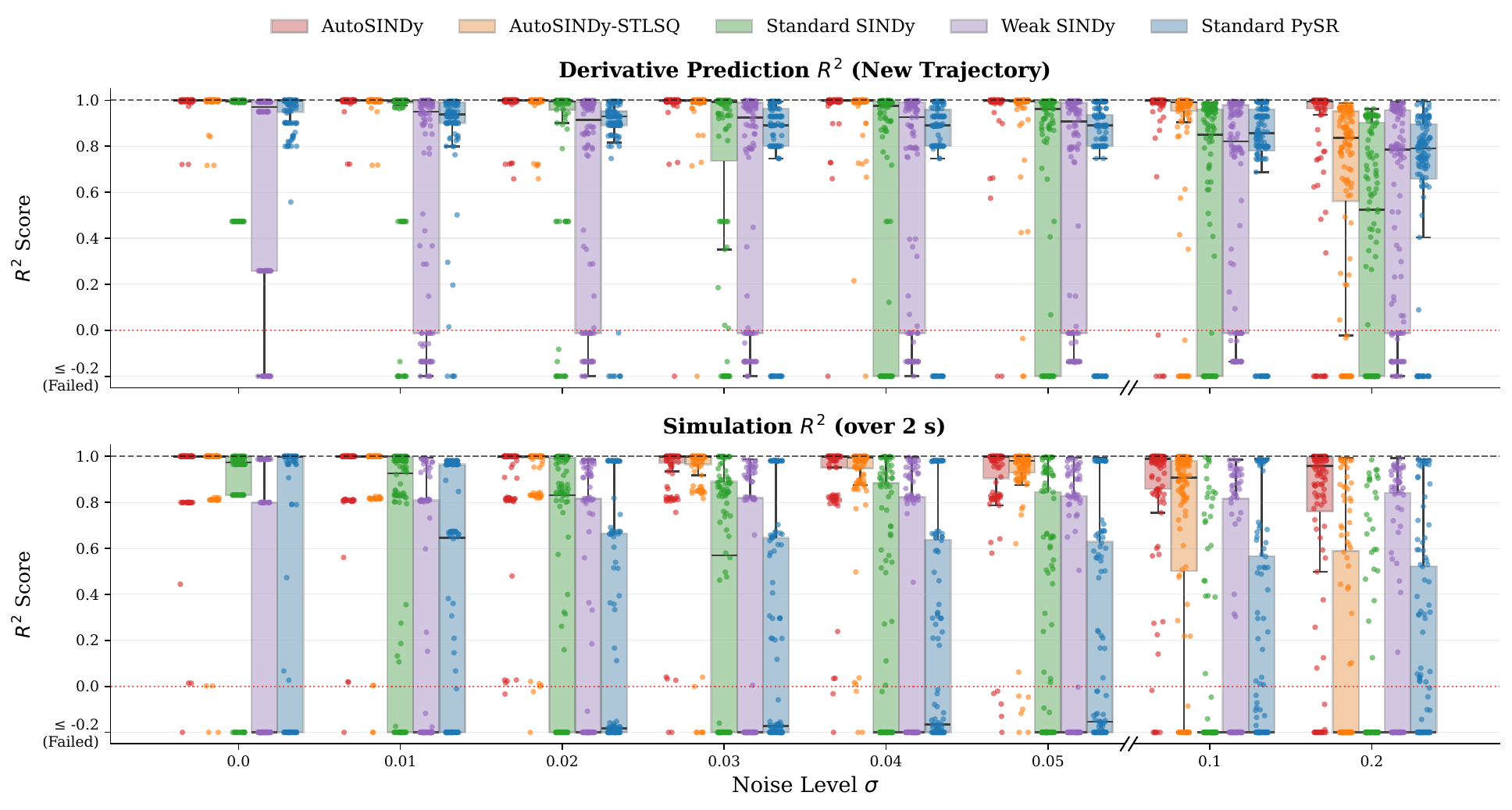}
  \caption{%
    Aggregate noise robustness distributions summarizing performance
    across all six systems. Boxplots are overlaid with raw trial data
    points. \textbf{Top:} Derivative prediction $R^2$ on the
    clean validation trajectory remains consistently near unity for
    \textsc{AutoSINDy} (red) across all noise levels. \textbf{Bottom:}
    Simulation $R^2$ shows \textsc{AutoSINDy} maintaining
    high performance at elevated noise while Standard SINDy (green) and
    Standard PySR (blue) exhibit increasing divergence rates. Trials with
    $R^2\leq-0.2$ \rev{are drawn at $-0.2$ and labelled}
    ``Failed''\rev{. The red dotted line marks $R^2=0$}.  \autosindy{} maintains the tightest, highest
    distribution\rev{, while Standard SINDy and Standard PySR} exhibit widespread failures,
    increasingly so with noise. \rev{The noise levels are evenly spaced on
    the horizontal axis, which is to scale only up to 0.05.}%
  }
  \label{fig:noise_robustness}
\end{figure*}

Figure~\ref{fig:noise_robustness} shows the distribution of
derivative-prediction and simulation $R^2$ scores aggregated across all
systems and plotted as a function of noise level $\sigma$.
\textsc{AutoSINDy}'s derivative-prediction distribution remains tightly
concentrated near $R^2=1$ for all noise levels, with only a small tail
of outlier trials (primarily on the Duffing system\rev{, and} at the
highest noise levels). In contrast, Standard SINDy exhibits a wide
variance that grows with noise, and Standard PySR's simulation
performance degrades at $\sigma=0.02$. Even at zero noise, the recovery rates reveal important structural
differences. Standard SINDy's \rev{83.3\%}, \rev{Standard} PySR's \rev{61.7\% and Weak SINDy's 50.0\%} confirm that
their failure modes are not purely noise-driven, they stem from
fundamental library and regularization limitations.  As noise grows,
\autosindy{} degrades most gracefully, maintaining \rev{91.7\%} recovery at
$\sigma=0.05$ \rev{and 67.5\% at $\sigma = 0.20$, compared to 35.8\% and 0\% for Standard SINDy, 25.0\% and 1.7\% for Weak SINDy, and 16.7\% and 1.7\% for Standard PySR. \autosindy-STLSQ reaches 83.3\% at $\sigma = 0.05$ but 0\% at $\sigma = 0.20$. Weak SINDy, which does not differentiate the noisy data, overtakes Standard SINDy only at high noise. Its recovery rate is higher from $\sigma = 0.10$ on, and its median derivative $R^2$ is higher at $\sigma = 0.20$ (0.79 against 0.52, Fig.~\ref{fig:noise_robustness}, top).}  Notably,
the derivative $R^2$ distribution of \autosindy{} in
Figure~\ref{fig:noise_robustness} remains tightly concentrated near
1.0 at all noise levels, which is a qualitatively different behaviour from the
wide, heavy-tailed distributions of \rev{the} baselines.

{\color{revcolor}
Table~\ref{tab:term_table} shows how well the methods find the true terms as
the noise grows. \autosindy{} has the highest mean TPR at every noise level,
from 0.87 without noise to 0.59 at $\sigma = 0.20$, while the best baseline,
Standard PySR, falls from 0.75 to 0.34.
}

\begin{table}[htbp]
\color{revcolor}
\centering
\caption{Term-level recovery by noise level. Each value is the mean TPR (Eq.~\eqref{eq:tpr}) over the 120 runs of one method at one noise level (six systems with 20 seeds each). The last row is the mean over all 960 runs of the method.}
\label{tab:term_table}
\renewcommand{\baselinestretch}{1}\small
\begin{tabular}{@{}lccccc@{}}
\toprule
 & \multicolumn{5}{c}{Mean TPR} \\
\cmidrule(l){2-6}
$\sigma$ & AutoSINDy & AutoSINDy-STLSQ & Standard SINDy & Weak SINDy & Standard PySR \\
\midrule
0.00 & 0.87 & 0.83 & 0.61 & 0.13 & 0.75 \\
0.01 & 0.86 & 0.83 & 0.39 & 0.13 & 0.61 \\
0.02 & 0.84 & 0.81 & 0.26 & 0.10 & 0.53 \\
0.03 & 0.81 & 0.75 & 0.21 & 0.11 & 0.44 \\
0.04 & 0.80 & 0.69 & 0.15 & 0.12 & 0.45 \\
0.05 & 0.77 & 0.63 & 0.13 & 0.11 & 0.41 \\
0.10 & 0.69 & 0.43 & 0.07 & 0.07 & 0.40 \\
0.20 & 0.59 & 0.25 & 0.05 & 0.05 & 0.34 \\
\midrule
All & 0.78 & 0.65 & 0.23 & 0.10 & 0.49 \\
\bottomrule
\end{tabular}
\end{table}

\begin{figure*}[t]
  \centering
  \includegraphics[width=\textwidth]{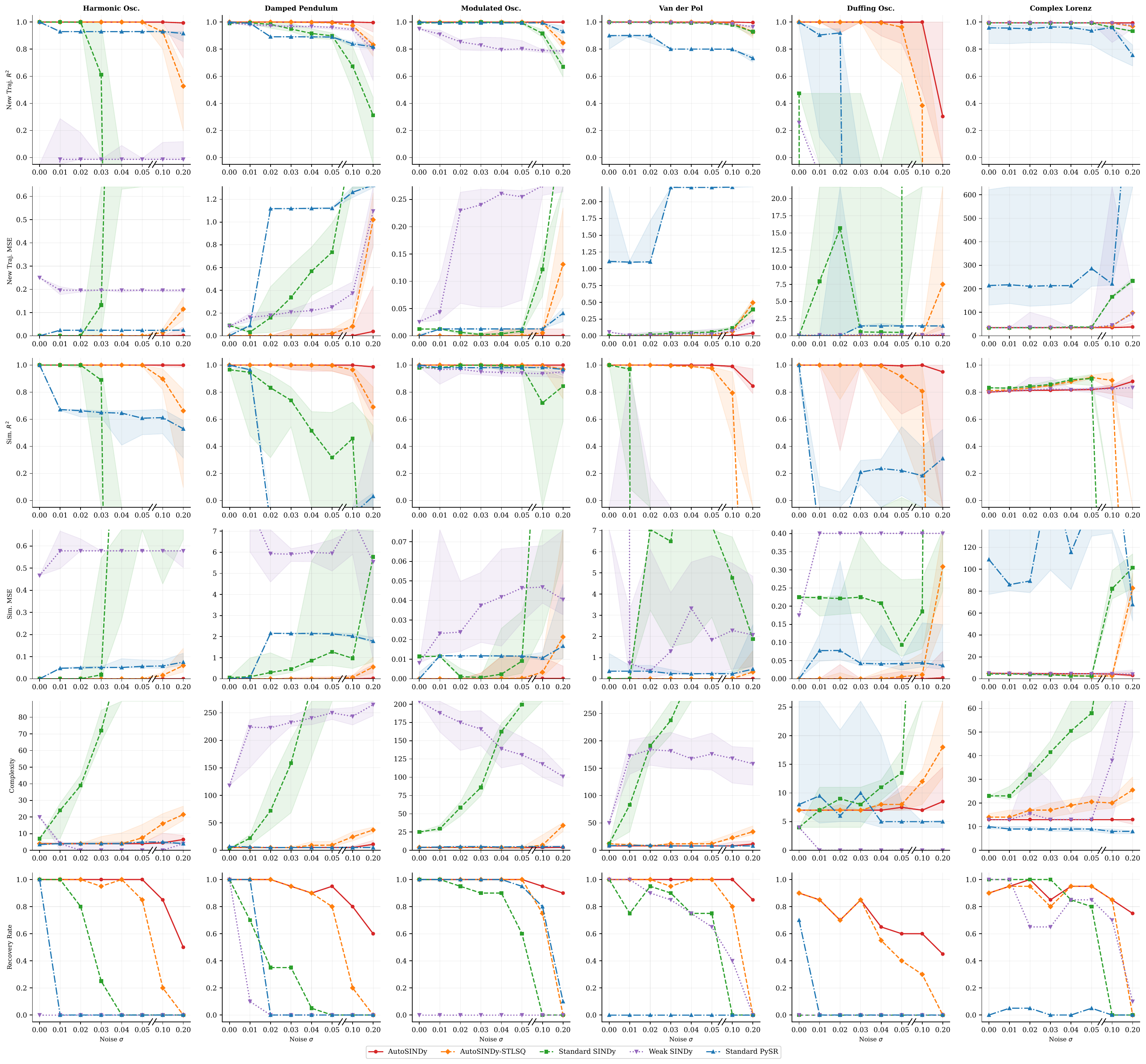}
  \caption{%
    Per-system performance dashboard. Each column corresponds to one of
    the six benchmark systems; each row shows one of six metrics as a
    function of noise level $\sigma$: validation-trajectory derivative
    $R^2$, derivative MSE, simulation $R^2$, simulation MSE, canonical
    complexity, and equation recovery rate ($R^2_{\mathrm{val}} \rev{>}
    0.99$). \rev{Lines} indicate the median\rev{, and} shaded bands denote the
    interquartile range over \rev{20} trials.
    \rev{Crashed simulations
    (Fig.~\ref{fig:sim_failures}) are left out of the simulation rows.
    The noise levels are evenly spaced on the horizontal axis, which is to
    scale only up to 0.05.}%
  }
  \label{fig:dashboard}
\end{figure*}

The per-system performance dashboard (Fig.~\ref{fig:dashboard}) reveals complementary system-specific patterns and the full noise-level breakdown for all six metrics per system. For the harmonic oscillator, \rev{the simplest of the six systems,} Standard SINDy experiences a striking validation failure. \rev{Its median validation $R^2$ is close to 1 up to $\sigma = 0.02$ but below zero from $\sigma = 0.04$ on.} This counterintuitive result is explained by severe library bloat: the \rev{median} canonical complexity reaches \rev{92} compared to the ground-truth value of 3. The enriched polynomial-Fourier basis contains high-degree cross-terms and trigonometric functions that are individually predictive on the noisy training data but mutually contradictory when fit simultaneously. \rev{In the STLSQ fit}, many of these spurious terms receive nonzero coefficients, producing an internally inconsistent model that fails on the clean validation trajectory. In contrast, \autosindy{} \rev{passes the recovery threshold} in every trial \rev{up to $\sigma = 0.05$ (91.9\% of all trials)} for this system.

This strong performance by \autosindy{} extends to other systems, \rev{passing the recovery threshold} in every trial \rev{up to $\sigma = 0.05$ (98.1\% of all trials)} for the modulated oscillator, whereas Standard SINDy again produces heavily over-complex equations (median \rev{128}) compared to the ground-truth complexity of \rev{5}. On the damped pendulum, which requires discovering the non-polynomial $\sin(x_0)$ term from noisy data, \autosindy{} achieves \rev{90.0\%} recovery \rev{(\autosindy-STLSQ 73.1\%)} compared to \rev{30.6\%} for Standard SINDy, \rev{13.8\% for Weak SINDy} and \rev{25.0\%} for Standard PySR. Standard SINDy also suffers from library bloat here, reaching a \rev{median complexity of 187} versus the true value of 4, leading to frequent simulation instabilities.

For the Van der Pol oscillator, \autosindy{} recovers a concise equation with a median complexity of \rev{8} and achieves \rev{near-}perfect simulation stability \rev{(no crashes, and one of its 160 simulations below $R^2 = 0$)}, again \rev{passing the recovery threshold} in every trial \rev{up to $\sigma = 0.10$ (98.1\% of all trials)}. Conversely, Standard SINDy's complexity explodes to a median of \rev{274} operators against a ground truth of 6, producing \rev{simulation crashes} in \rev{31 of 160} trials (see Section~\ref{subsec:results_sim}). Furthermore, Standard PySR achieves 0\% recovery on this system despite producing a median $R^2=0.8$, which consistently falls below the $R^2>0.99$ threshold. This confirms that pure symbolic regression, without sparsity promotion, \rev{often} fails to isolate the \rev{linear term $-x_0$ of $\dot{x}_1$ (it is missing in 116 of 160 runs)}.

The Duffing oscillator proves to be the most challenging case for all methods. \autosindy{} achieves \rev{70.0\%} recovery \rev{(\autosindy-STLSQ 56.9\%)}, \rev{clearly} outperforming the 0\% for Standard SINDy \rev{and Weak SINDy} and \rev{8.8\%} for Standard PySR. The bimodal distribution of outcomes, where trials either succeed completely ($R^2\approx1$) or fail catastrophically, is visible in Figure~\ref{fig:dashboard} as a large interquartile range shaded region for \autosindy{} on the Duffing column. While Standard SINDy and Standard PySR \rev{crash} in 8 and \rev{16 of 160} trials respectively, \autosindy{} \rev{crashes in only 2, both at $\sigma \geq 0.10$} (see Section~\ref{subsec:results_sim}).

Finally, the Complex Lorenz system represents the most structurally challenging case, featuring a 3-dimensional chaotic attractor and a mixed polynomial-trigonometric coupling term. All methods show higher variance here, yet \autosindy{} \rev{has no simulation crashes (6 of its 160 simulations score below zero)} and a median complexity of \rev{13}, closely matching the ground truth of 15. Standard SINDy attains \rev{the same median derivative $R^2$ (0.994) and a marginally higher median simulation $R^2$ of 0.832 versus 0.813} for \autosindy{}, but this comes at the cost of \rev{3.5}$\times$ the model complexity (see Section~\ref{subsec:results_complexity}). \rev{The medians over all noise levels hide a change with noise, however. Standard SINDy passes the recovery threshold in every trial up to $\sigma = 0.03$ but in none at $\sigma = 0.10$ and $0.20$, where its median complexity grows to 109.5 and 152.5, while \autosindy{} still passes in 85\% and 75\% of the trials (Fig.~\ref{fig:dashboard}).}

Full per-trial trajectory comparisons, state simulations, and phase
portraits for all six systems are provided in Appendix~\ref{app:per_system_results}
(Figs.~\ref{fig:app_harmonic}--\ref{fig:app_lorenz}).
 
\subsection{Simulation Failure Analysis}
\label{subsec:results_sim}
 
\begin{figure*}[t]
  \centering
  \includegraphics[width=\textwidth]{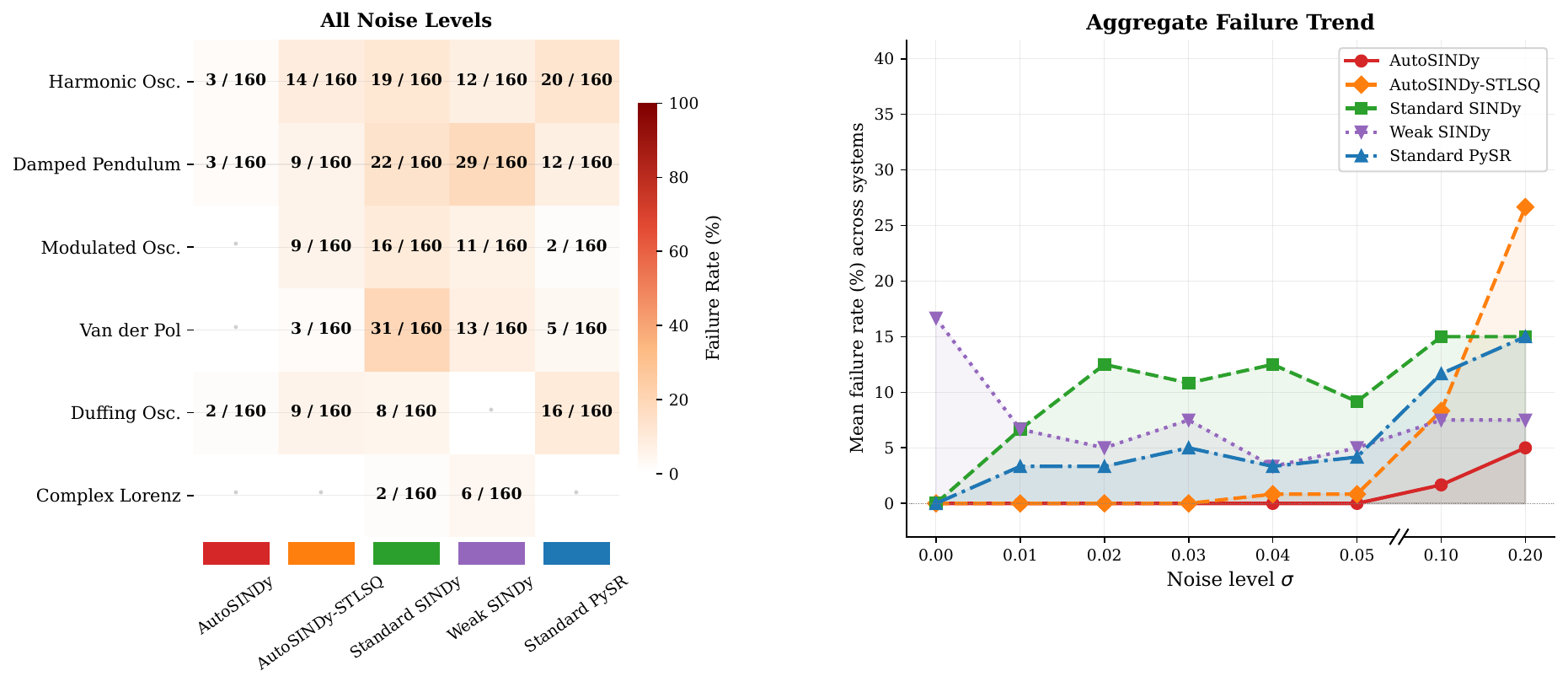}
  \caption{Simulation failure analysis.
    \textbf{Left:} Heatmap of integration failure counts (out of \rev{160}
    trials) per system and method.  \autosindy{} produces \rev{8 hard
    integration crashes in total, all at $\sigma = 0.10$ and $0.20$. Standard
    SINDy failures are concentrated on Van der Pol (31/160) and the damped
    pendulum (22/160), and Standard PySR failures on the harmonic oscillator
    (20/160) and the Duffing oscillator (16/160).}
    \textbf{Right:} Mean failure rate trend vs.\ noise level,
    aggregated across systems.  \autosindy{} (red) remains at 0\%
    \rev{up to $\sigma = 0.05$ and reaches 5\% at $\sigma = 0.20$}, while
    Standard SINDy (green) and Standard PySR (blue) \rev{reach 15\%. The
    noise levels are evenly spaced on the horizontal axis, which is to scale
    only up to 0.05.}}
  \label{fig:sim_failures}
\end{figure*}

Figure~\ref{fig:sim_failures} presents the simulation failure
analysis.  The left heatmap shows hard integration failures (NaN,
Inf, or early termination due to the divergence safety brake \rev{or the time limit}) per
system.  The right panel shows the aggregate mean failure rate trend
as noise increases.
 
A critical finding from Figure~\ref{fig:sim_failures} is that
\autosindy{} produces \rev{the fewest} hard simulation crashes\rev{. Of 960
trials per method, \autosindy{} crashed in 8, all at $\sigma \geq 0.10$, and
the other methods in 44 to 98. Runs that did not crash but scored below $-1$
numbered 11 for \autosindy{} and 67 to 534 for the other methods.} Standard SINDy produces hard failures \rev{most often on
Van der Pol (31/160) and the damped pendulum (22/160)}, and
Standard PySR \rev{on the harmonic oscillator (20/160) and the Duffing
oscillator (16/160)}.

The right panel of Figure~\ref{fig:sim_failures} shows the aggregate
failure trend with noise.  While \autosindy{} maintains a \rev{0\%
crash rate up to $\sigma = 0.05$ and reaches 1.7\% at $\sigma = 0.10$ and
5.0\% at $\sigma = 0.20$}, Standard SINDy \rev{crashes in 6.7--12.5\% of the
trials from $\sigma = 0.01$ to $0.05$ and in 15.0\% at $\sigma = 0.10$ and
$0.20$}. Standard PySR \rev{crashes in 3.3--5.0\% of the trials from
$\sigma = 0.01$ to $0.05$ and rises to 15.0\% at $\sigma = 0.20$.
\autosindy-STLSQ has no crashes up to $\sigma = 0.03$ but the highest rate
at $\sigma = 0.20$ (26.7\%), and Weak SINDy crashes most often on
noise-free data (16.7\%).}

The distinction between hard crashes (Figure~\ref{fig:sim_failures})
and simulation-tier failures (Figure~\ref{fig:tiers}) is
important\rev{. Most simulations in the Failed tier ($R^2<0$) did not crash.
The integration finished, but the trajectory was physically incorrect.
This holds for every method. For \autosindy{}, 30 of its 38 Failed trials
did not crash, and for Standard SINDy, 381 of 479.}

\subsection{Model Parsimony}
\label{subsec:results_complexity}
 
\begin{figure}[t]
  \centering
  \includegraphics[width=\columnwidth]{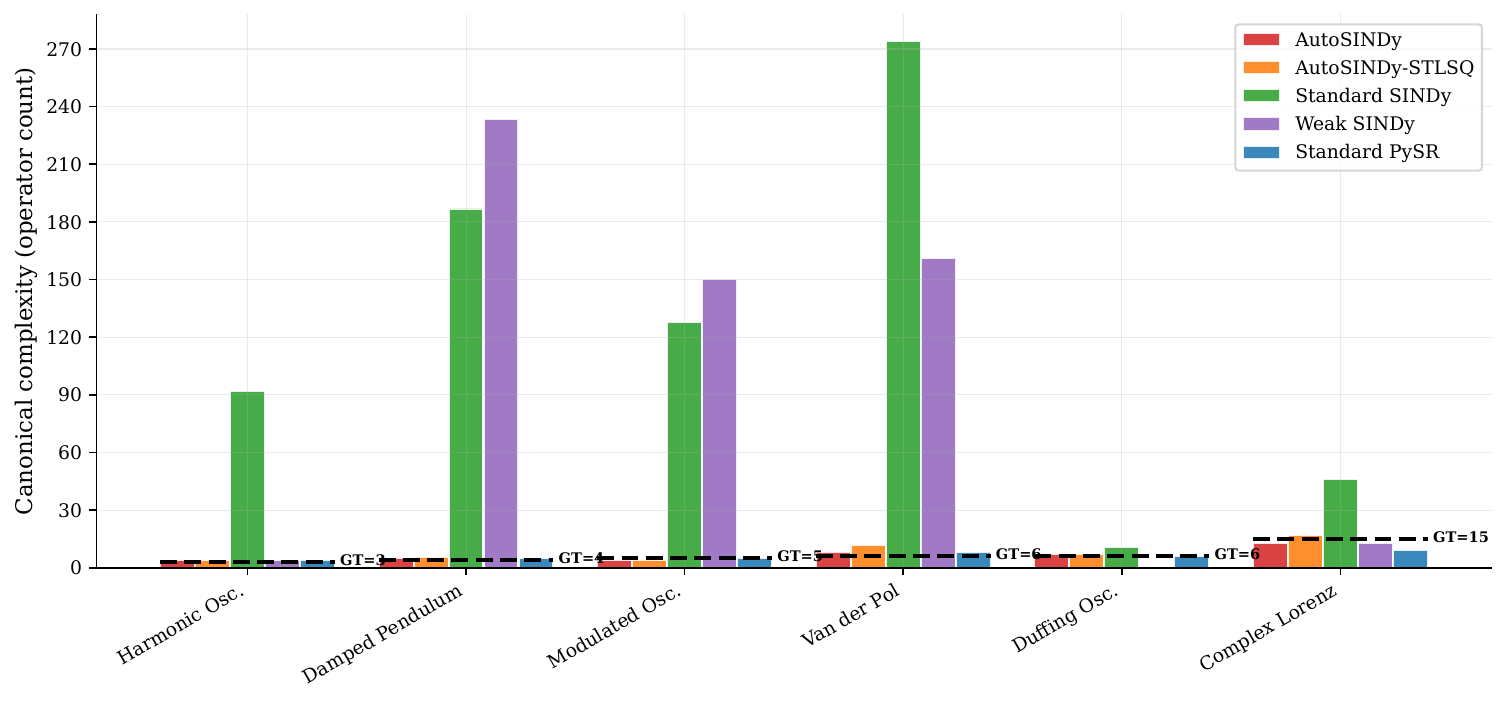}
  \caption{%
    Model complexity (median canonical operator count across all noise
    levels and trials). Dashed lines mark the ground-truth complexity of
    each system. \textsc{AutoSINDy} (red) recovers near-ground-truth
    complexity on every system. Standard SINDy (green) is severely
    over-complex due to its large fixed polynomial-Fourier library.
    Standard PySR (blue) \rev{stays close to the ground-truth complexity
    except on the Complex Lorenz system, where it is under-complex (median 9
    against 15)}, indicating that
    standalone symbolic regression compresses equations at the cost of
    accuracy.%
  }
  \label{fig:complexity}
\end{figure}

Figure~\ref{fig:complexity} compares the median canonical complexity of
the discovered equations against the ground-truth values (dashed line)
for each system. \textsc{AutoSINDy}
recovers equations within 1--2 operators of the ground-truth complexity
on all six systems. Standard SINDy's enriched library induces dramatic
over-complexity: \rev{30.7}$\times$ on Harmonic Oscillator, \rev{46.8}$\times$ on
Damped Pendulum, \rev{45.7}$\times$ on Van der Pol, and \rev{25.6}$\times$ on
Modulated Oscillator. The Van der Pol case is particularly striking:
the true equation requires only 6 operations, but Standard SINDy
discovers equations with a median of \rev{274} operations, showing that
\rev{such a model}, while fitting the \rev{derivatives} well, is far too complex to simulate
stably and carries no interpretable physical meaning. Standard PySR,
by contrast, \rev{finds compact models, close to the ground-truth
complexity except on the Complex Lorenz system (median 9 against 15)},
at the cost of missing critical
terms, which explains its superior parsimony scores alongside its poor
simulation stability.
 
\begin{figure}[t]
  \centering
  \includegraphics[width=\columnwidth]{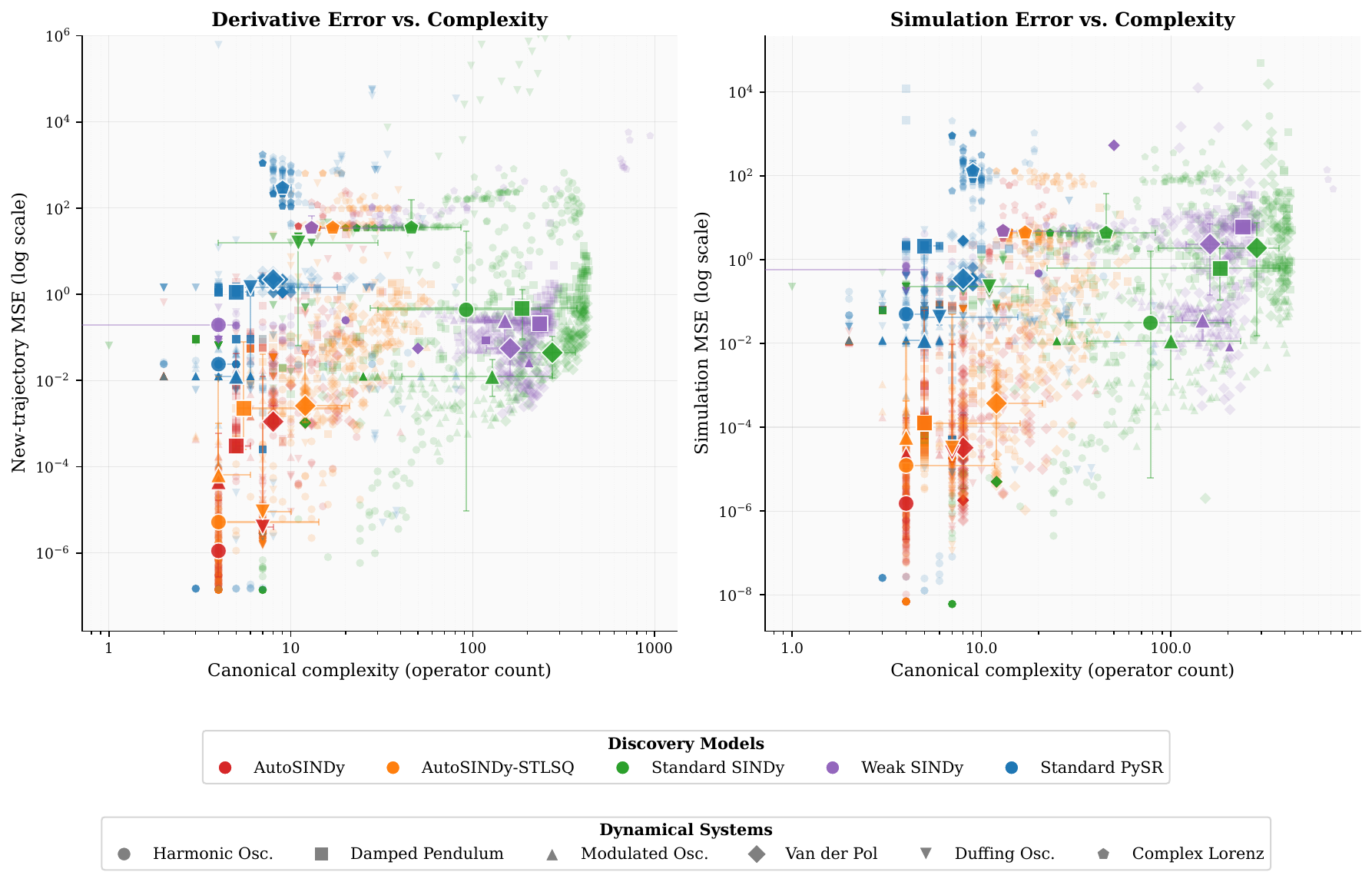}
  \caption{%
Scatter plots comparing canonical model complexity against prediction error (Derivative MSE and Simulation MSE) on a logarithmic scale. Both panels use shared color encodings for the discovery algorithms and distinct marker shapes for the underlying dynamical systems. Solid markers with error bars denote the median performance and interquartile ranges, while faded background markers represent individual trial runs.
    \textsc{AutoSINDy} consistently occupies the lower-left
    region, combining low complexity with low error. \rev{The vertical axis
    of the left panel ends at $10^6$, so the few runs with larger errors are
    not shown.}%
  }
  \label{fig:scatter}
\end{figure}

The error-vs.-complexity scatter plots (\rev{Fig.}~\ref{fig:scatter}) confirm this picture across all systems
simultaneously. \textsc{AutoSINDy}'s trials cluster in the
lower-left region of both plots, which means low complexity \emph{and} low
error, while Standard SINDy's trials span a wide range of high
complexities with no corresponding accuracy benefit, and Standard PySR's
trials show low complexity but substantially higher error.
 
\subsection{Computational Efficiency}
 
\begin{figure}[t]
  \centering
  \includegraphics[width=\columnwidth]{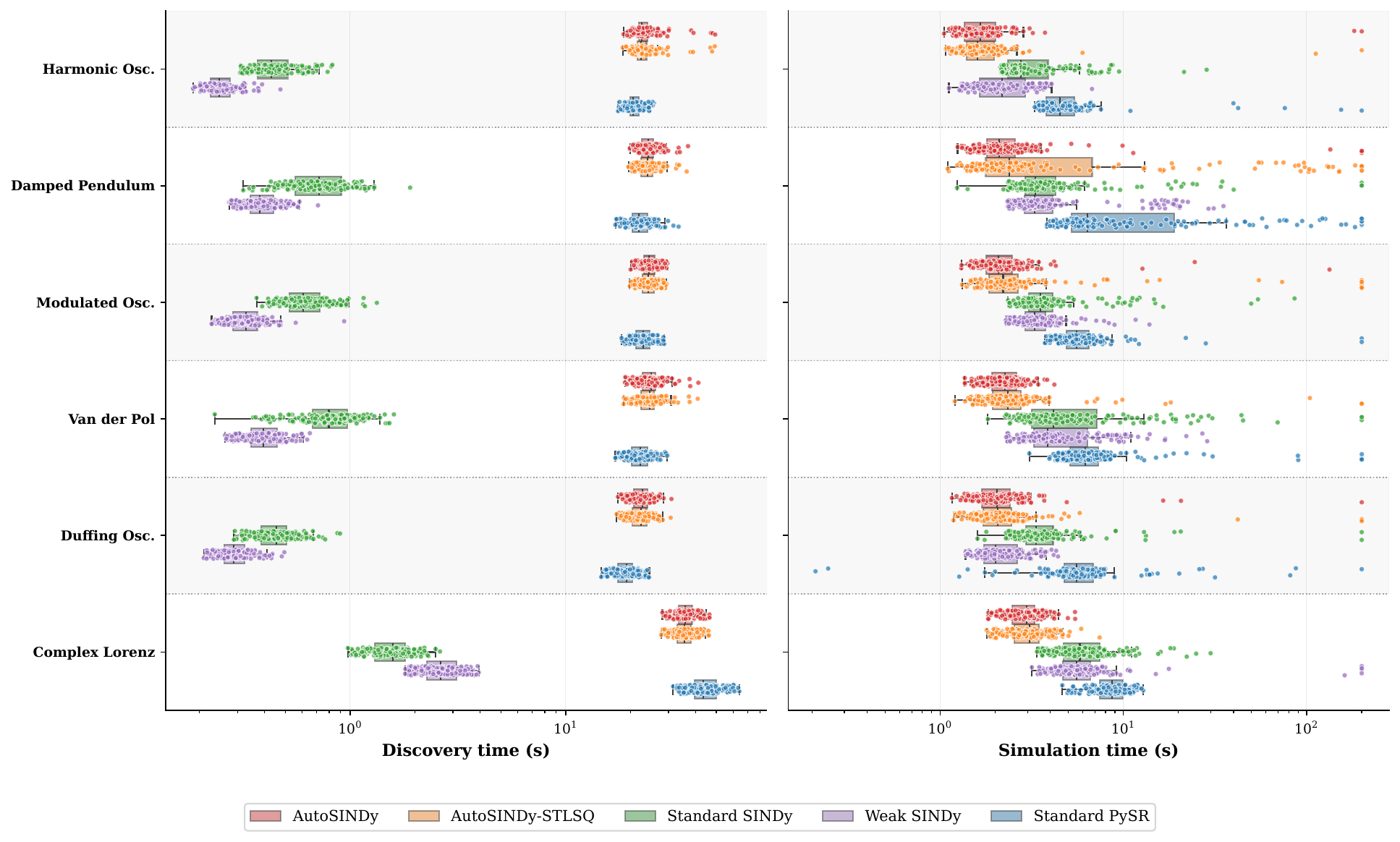}
  \caption{%
    Computation time distributions for discovery (left) and simulation
    (right). Both axes use logarithmic scale. \textsc{AutoSINDy}'s
    discovery time (\rev{median 24\,s}) is comparable to or slightly higher
    than standalone Standard PySR, and \rev{about 37 times} longer
    than Standard SINDy, which does not require symbolic search. Simulation
    times for \textsc{AutoSINDy} are uniformly short and low-variance
    due to the compactness of the discovered equations; Standard SINDy
    and Standard PySR exhibit higher and more variable simulation times.
    \rev{The times were measured with three runs in parallel on one laptop,
    and points at 200\,s are simulations stopped at the 200\,s limit.}%
  }
  \label{fig:timing}
\end{figure}

Figure~\ref{fig:timing} shows wall-clock discovery and simulation times
as boxplots per system. \autosindy{} and Standard PySR incur comparable discovery overheads
($\sim$20--40~s, median $\sim$24~s and $\sim$\rev{22}~s respectively)
reflecting that both invoke the same \pysr{} engine.  Standard SINDy
requires \rev{a median of 0.65~s} for library fitting.  The discovery overhead of
\autosindy{} is therefore the cost of automation: it replaces the
manual feature-engineering step with a systematic search.
 
Importantly, \autosindy{} is consistently \rev{among} the \emph{fastest} \rev{methods}
for simulation across all systems, with a tight distribution
concentrated below \rev{4}~s (median $\sim$\rev{2.2}~s).  Standard SINDy's
simulation time is highly variable, including outliers exceeding
$10^2$~s, driven by numerically stiff ODE systems arising from its
bloated coefficient structures.  Standard PySR shows the highest
and most variable simulation times (\rev{median $\sim$6~s}, with extreme
outliers), consistent with frequent
near-divergent trajectories that require small integrator steps.
 
The compact equations recovered by \autosindy{} therefore provide a
compounding advantage: they are not only more interpretable but also
substantially cheaper to simulate, which is a property of direct practical
value in real-time or embedded applications.
 
\subsection{Summary of Quantitative Results}

Table~\ref{tab:system_master_summary} presents a comprehensive performance breakdown of AutoSINDy compared to standard baseline methods across six canonical dynamical systems. To evaluate robustness against observational noise, we report the median metrics (and mean Recovery Rate \rev{and TPR}) aggregated across all noise levels and random seeds. \rev{Figure~\ref{fig:dashboard} shows $R^2$ and MSE for each noise level.} \rev{Generalization to a different initial condition is measured on the separate validation trajectory.} The evaluation encompasses structural discovery accuracy (Derivative \rev{NRMSE}), integration stability (Simulation \rev{NRMSE}), model parsimony (Canonical Complexity), the overall equation Recovery Rate\rev{, and term-level recovery (TPR)}.

As demonstrated in Table~\ref{tab:system_master_summary}, AutoSINDy consistently outperforms the baselines, maintaining near-perfect predictive accuracy (Derivative \rev{NRMSE at most 0.010}) and stable forward simulation (Simulation \rev{NRMSE at most 0.023}) across the majority of the evaluated systems. Furthermore, AutoSINDy successfully avoids equation bloat, consistently discovering parsimonious models whose canonical complexity perfectly or closely matches the ground truth.

In contrast, Standard SINDy frequently suffers from catastrophic simulation divergence (evidenced by \rev{a median Simulation NRMSE above 1, that is $R^2 < 0$, on Van der Pol and the Duffing oscillator}), even in cases where it achieves reasonable derivative approximations. Meanwhile, Standard PySR struggles significantly with both forward simulation stability and structural recovery, often yielding low recovery rates. These results underscore the efficacy of AutoSINDy's hybrid discovery-then-solve framework in identifying equations that are not only structurally accurate but also dynamically stable \rev{in simulation}.

\begin{table*}[htbp]
\centering
\caption{Aggregated Performance by System. \rev{Each value is the median over the 160 runs of one method on one system (eight noise levels with 20 seeds each). The recovery rate and the TPR are means.} Bold \rev{marks the best value in each column for each system, and tied values are all bold. For complexity, bold marks the value closest to the ground truth.}}
\label{tab:system_master_summary}
\renewcommand{\baselinestretch}{1}\small
\color{revcolor}
\begin{tabular}{@{}llccccc@{}}
\toprule
\textbf{System} & \textbf{Model} & Deriv. NRMSE & Sim. NRMSE & Comp. & Recov. & TPR \\
\midrule
\multirow{5}{*}{Harmonic Osc.} & \textbf{AutoSINDy} & \textbf{0.003} & \textbf{0.003} & \textbf{4.0} & \textbf{91.9\%} & \textbf{0.89} \\
 & AutoSINDy-STLSQ & 0.007 & 0.011 & \textbf{4.0} & 75.0\% & 0.72 \\
 & Standard SINDy & 1.131 & 0.926 & 92.0 & 38.1\% & 0.27 \\
 & Weak SINDy & 1.006 & 1.953 & \textbf{4.0} & 0.0\% & 0.00 \\
 & Standard PySR & 0.266 & 0.592 & \textbf{4.0} & 12.5\% & 0.53 \\
\midrule
\multirow{5}{*}{Damped Pendulum} & \textbf{AutoSINDy} & \textbf{0.005} & \textbf{0.008} & \textbf{5.0} & \textbf{90.0\%} & \textbf{0.82} \\
 & AutoSINDy-STLSQ & 0.016 & 0.017 & 5.5 & 73.1\% & 0.67 \\
 & Standard SINDy & 0.254 & 0.676 & 187.0 & 30.6\% & 0.16 \\
 & Weak SINDy & 0.181 & 1.782 & 233.5 & 13.8\% & 0.00 \\
 & Standard PySR & 0.330 & 1.076 & \textbf{5.0} & 25.0\% & 0.34 \\
\midrule
\multirow{5}{*}{Modulated Osc.} & \textbf{AutoSINDy} & \textbf{0.005} & \textbf{0.005} & 4.0 & \textbf{98.1\%} & 0.51 \\
 & AutoSINDy-STLSQ & 0.006 & 0.010 & 4.0 & 84.4\% & 0.44 \\
 & Standard SINDy & 0.074 & 0.137 & 128.0 & 66.9\% & 0.10 \\
 & Weak SINDy & 0.393 & 0.224 & 150.5 & 0.0\% & 0.00 \\
 & Standard PySR & 0.075 & 0.139 & \textbf{5.0} & 85.6\% & \textbf{0.57} \\
\midrule
\multirow{5}{*}{Van der Pol} & \textbf{AutoSINDy} & \textbf{0.010} & \textbf{0.023} & \textbf{8.0} & \textbf{98.1\%} & \textbf{0.91} \\
 & AutoSINDy-STLSQ & 0.016 & 0.076 & 12.0 & 84.4\% & 0.70 \\
 & Standard SINDy & 0.078 & 1.611 & 274.0 & 63.8\% & 0.17 \\
 & Weak SINDy & 0.071 & 5.253 & 161.0 & 69.4\% & 0.00 \\
 & Standard PySR & 0.447 & 2.201 & \textbf{8.0} & 0.0\% & 0.60 \\
\midrule
\multirow{5}{*}{Duffing Osc.} & \textbf{AutoSINDy} & \textbf{0.006} & \textbf{0.020} & 7.0 & \textbf{70.0\%} & \textbf{0.81} \\
 & AutoSINDy-STLSQ & 0.009 & 0.027 & 7.0 & 56.9\% & 0.75 \\
 & Standard SINDy & 11.303 & 1.877 & 11.0 & 0.0\% & 0.20 \\
 & Weak SINDy & 1.066 & 2.324 & 0.0 & 0.0\% & 0.00 \\
 & Standard PySR & 3.425 & 0.892 & \textbf{6.0} & 8.8\% & 0.46 \\
\midrule
\multirow{5}{*}{Complex Lorenz} & \textbf{AutoSINDy} & \textbf{0.076} & 0.433 & \textbf{13.0} & \textbf{90.0\%} & \textbf{0.74} \\
 & AutoSINDy-STLSQ & 0.077 & 0.413 & \textbf{17.0} & 79.4\% & 0.63 \\
 & Standard SINDy & 0.078 & \textbf{0.410} & 46.0 & 70.6\% & 0.48 \\
 & Weak SINDy & \textbf{0.076} & 0.434 & \textbf{13.0} & 72.5\% & 0.61 \\
 & Standard PySR & 0.243 & 2.197 & 9.0 & 1.9\% & 0.45 \\
\bottomrule
\end{tabular}
\end{table*}

\section{\label{sec:discussion}Discussion}
 
\subsection{Why \autosindy{} Outperforms Its Components in Isolation}

The performance gap between \autosindy{} and its two individual
components, \pysr{} alone and \sindy{} alone, is instructive.
Standard PySR achieves \rev{22.3\%} Excellent derivative
performance (Figure~\ref{fig:tiers}) and collapses to
\rev{6.6\%} in simulation.  This discrepancy traces to the lack of
regularization\rev{. Without} a sparsity-promoting mechanism, genetic
programming optimizes fit on \rev{the noisy training data}, where even
slightly incorrect coefficient values \rev{or a missing term} produce
models that diverge \rev{in simulation}.  Particularly telling is the 0\% recovery
on the Van~der~Pol oscillator\rev{. PySR usually} identifies $x_0^2
x_1$ as a relevant term, but \rev{it misses the linear term $-x_0$ in
116 of 160 runs (Sec.~\ref{subsec:results_noise}), an error} that
accumulates into trajectory divergence.

Standard SINDy suffers from two distinct failure modes
depending on the relationship between its fixed library and the true
governing structure.  The first is over-completeness under
noise\rev{. The} polynomial-Fourier basis contains the correct terms for
most systems (e.g., $x_0^2 x_1$ for Van~der~Pol, $\sin(x_0)$ for
the damped pendulum), but also \rev{dozens of} correlated alternatives
\rev{(50 terms in all for two state variables)}.
Under noisy derivative estimates, the optimizer distributes
coefficients across these spurious terms rather than concentrating
them on the few correct ones, producing the catastrophic equation
bloat visible in Figure~\ref{fig:complexity}\rev{, with 92} operators for
the harmonic oscillator against a ground truth of 3, and \rev{187} for the
damped pendulum against a ground truth of 4.  The second failure mode
is genuine structural incompleteness\rev{. The Fourier library}
applies $\sin/\cos$ to individual state variables only, so the
compound term $x_1\sin(x_0+x_2)$ in the Complex Lorenz system is
genuinely absent.  SINDy approximates it with many polynomial-Fourier
cross-products, explaining simultaneously its \rev{70.6\%}
recovery rate and its \rev{46}-operator complexity against a ground truth
of 15. \rev{A polynomial library can approximate the $\sin x_0$ term of the
damped pendulum through its Taylor series. On a larger swing than any in the
training data, however, the resulting models diverge or lose accuracy, whereas
\autosindy-STLSQ, with the same optimizer, finds $\sin x_0$ itself
(Appendix~\ref{app:pendulum_degrees}).}

\autosindy{} \rev{addresses} both failure modes by construction.
\rev{Stage~1} actively mines for the system-specific
functional forms present in the data, rather than assuming they belong
to a fixed class.  \rev{Stage~2} then eliminates
over-completeness by pruning the discovered pool to a compact,
non-redundant basis before any regression is performed.
 
\subsection{The Role of Library Curation}

The curation stage is not merely a preprocessing convenience\rev{. It keeps
the library small and its terms distinct.}  Without
collinearity pruning, $\boldsymbol{\Theta}$ can be near-singular. \rev{When
PySR returns two nearly identical candidates, the sparse regression must
choose between them.}  By rejecting any candidate whose correlation with
already-accepted terms exceeds a threshold, the curation
pipeline \rev{keeps only one term of each such pair.}

The simplicity bias, which is sorting candidates in ascending order of SymPy
operator count before applying the correlation filter, is what makes
this pruning meaningful rather than arbitrary.  By always admitting
the simplest expression that captures a given piece of information,
the curation algorithm directly implements a form of Occam's razor.
Complex terms are rejected not because they are wrong, but because a
simpler term already explains the same variance in the data.
Therefore, the downstream \rev{sparse regression} can then concentrate all
coefficient mass on the small number of genuinely active terms.
 
\subsection{Simulation Stability as a Test of Physical Validity}
 
The \rev{low} hard-crash rate of \autosindy{} \rev{(8 of 960 trials, the
fewest of the five methods, all at $\sigma \geq 0.10$)}
(Figure~\ref{fig:sim_failures}) is a substantive finding rather than
a secondary observation.  An integration crash, where state variables
diverge to infinity within the simulation window, indicates that the
discovered model assigns arbitrarily large derivatives to reachable
states.  This is not merely numerical inaccuracy\rev{. It} reflects a
physically incoherent equation structure in which the right-hand side
of the ODE grows without bound.  Standard SINDy produces hard crashes
in \rev{98 of its 960 trials, most often on Van~der~Pol (31/160) and the
damped pendulum (22/160)}. Standard PySR crashes in \rev{55, most often on
the harmonic oscillator (20/160) and the Duffing oscillator (16/160)}.

The aggregate failure trend in Figure~\ref{fig:sim_failures} (right
panel) reveals an important distinction\rev{. Neither Standard SINDy nor
Standard PySR crashes on noise-free data, but Standard SINDy crashes in
6.7\% of the trials already at $\sigma = 0.01$, whereas Standard PySR
stays at or below 5.0\% up to $\sigma = 0.05$.}  SINDy crashes are induced by
noise inflating the derivative estimates and causing the over-complete
optimizer to assign destabilizing coefficients to high-degree
terms\rev{.} \autosindy's \rev{avoidance of hard crashes up to $\sigma = 0.05$
on} all six systems confirms that the discovered equations, whatever their
coefficient errors, remain within a plausible physical manifold.

\subsection{The Duffing Oscillator: Limits of Noise Robustness}
\label{subsec:duffing_discussion}
 
The Duffing oscillator ($\dot{x}_1=-\delta x_1-\alpha x_0-\beta
x_0^3$) exposes a fundamental boundary of the current framework.
Its bistable potential ($\alpha<0$, $\beta>0$) \rev{has two wells, and
the training data stay in one of them}.  \rev{Across all noise levels},
all \rev{methods} struggle\rev{.} \autosindy{} achieves 70\rev{.0}\%
recovery (vs.\ 0\% for Standard SINDy \rev{and Weak SINDy} and \rev{8.8}\% for
\rev{Standard} PySR).  The
per-system dashboard (Figure~\ref{fig:dashboard}, Duffing
column) shows that \autosindy's \rev{median} simulation $R^2$ \rev{stays
above 0.99 up to $\sigma = 0.10$. Which part of the motion the training
data cover matters more than how many points they contain. When we
trained on only the middle 50\% or 65\% of the recording, which miss the
first large swing, no method reached a median $R^2$ above 0.84 on the
new trajectory. With the first swing included, \autosindy{} reached
0.97--1.000, whereas no baseline reached 0.95 with any of these windows
(Appendix~\ref{app:settings}).}
 
\subsection{Parsimony, Interpretability, and Downstream Scientific Value}
 
Figure~\ref{fig:complexity} demonstrates that \autosindy{}
consistently produces equations within a narrow margin of the true
ground-truth complexity.  This parsimony is not merely an aesthetic
preference\rev{. It} has concrete practical consequences.

Discovered equations at \rev{1.13}$\times$ the ground-truth complexity can
be directly submitted to downstream scientific analysis.  For a
dynamical system described by 4 or 6 operators, an analyst can compute
equilibria analytically, perform linearization and stability analysis,
derive conservation laws, apply perturbation theory, or use the
equation as a mechanistic prior in a downstream Bayesian model.  None
of these operations is tractable on a Van~der~Pol model with \rev{274}
operators or a harmonic oscillator model with \rev{92} operators, even if
their derivative prediction scores are nominally acceptable.
 
The timing results (Figure~\ref{fig:timing}) make the same point in
computational terms.  \autosindy{} is systematically \rev{among} the fastest
\rev{methods} for simulation. This is not because of any algorithmic optimization of
the integrator, but because its discovered equations have fewer
right-hand-side operations per step, producing less stiff ODE systems
that the Radau solver can traverse with larger time steps.  Standard
SINDy's bloated equations, by contrast, frequently produce ODE systems
whose stiffness requires extremely fine time-stepping, explaining the
outliers reaching $10^2$~s in Figure~\ref{fig:timing} (right column)
even on two-dimensional systems like the \rev{damped pendulum and the
Van~der~Pol oscillator}.
 
The contrast between derivative $R^2$ and simulation speed therefore
captures a key insight that the prediction accuracy on observed states is a
necessary but insufficient criterion for equation quality.  A concise,
physically coherent equation is simultaneously more accurate, more
stable, more interpretable, and faster to simulate than a bloated
approximation with a comparable short-horizon fit.  \autosindy{}
demonstrates that these desiderata can be jointly optimized through
principled basis construction rather than requiring a post-hoc
trade-off.
 
\subsection{Limitations and Future Work}
 
Several limitations of the current framework merit acknowledgment.

{\color{revcolor}
\paragraph{Terms the data cannot tell apart.}
The curation in Stage~2 keeps the simpler of two terms that are almost
proportional on the training data, so a true term is lost when a simpler
term looks almost the same. On the modulated oscillator, the true damping
term $x_1\cos x_0$ is almost proportional to $x_1$ on the small swings of
the training data, so Stage~2 keeps $x_1$. In 118 of the 160 runs on this
system, \autosindy's model contains $x_1$ in place of $x_1\cos x_0$, and
117 of these models still reach $R^2 > 0.99$ on the new trajectory. Data
with larger swings, on which the two terms differ more, would be needed to
separate them.

\paragraph{Derivative estimation in Stage~1.}
Stage~3 of \autosindy{} works without estimated derivatives, but in
Stage~1, \pysr{} learns from derivatives estimated with \pysindy's default
Savitzky--Golay filter, the same filter as for the other methods
(Sec.~\ref{subsec:hyperparams}). With very noisy data, users are encouraged
to try stronger smoothing or other derivative estimators~\cite{Kaptanoglu2022}.
Running \pysr{} on local averages of the data, which needs no derivatives,
could also make \autosindy{} even more robust to noise.
}

\paragraph{Computational cost.}
\textsc{AutoSINDy}'s discovery time (\rev{$10$--$14$}\,s per state variable)
is moderate for two- and three-dimensional systems but scales linearly
in the number of state variables and chunks, which could become
prohibitive for high-dimensional systems. Promising mitigation
strategies include: amortized symbolic regression (pre-training a
surrogate to predict candidate terms from data statistics), parallel
\pysr{} calls per state variable, and warm-starting from libraries
curated on related systems.

\paragraph{\rev{Simulation of chaotic systems.}}
The Complex Lorenz simulation results confirm that \rev{an accurate fit
of the derivatives} does not guarantee accurate
simulation for chaotic systems.  Chaotic sensitivity means any
integration error compounds exponentially; even the true model with
finite-precision arithmetic will diverge from a specific reference
trajectory.  Future evaluation should use attractor-geometry metrics that are
insensitive to phase shifts.

{\color{revcolor}
\paragraph{Other last stages.}
\autosindy{} and \autosindy-STLSQ use the same symbolic mining and library
construction but two different last stages. Bayesian versions of
\sindy{}~\cite{Hirsh2022,Fung2025} and methods that estimate the noise-free
states together with the model~\cite{Kaheman2022,Hokanson2023} are natural
further candidates for the last stage.
}

\paragraph{Extension to PDEs and real data.}
All experiments use synthetic ODE data with controlled Gaussian noise
\rev{of up to $\sigma = 0.20$}. \rev{Natural next test cases are
higher-dimensional systems, colored noise and process noise (stochastic
differential equations).}
Extension to partial differential equations would require spatially
local symbolic regression on data slices, followed by global library
curation enforcing spatial consistency.  Applying \autosindy{} to
real experimental time series, where the ground truth is unknown, data may be irregular, and
state variables may be partially observed, remains a high-priority
validation direction and a natural next step given the framework's
noise robustness demonstrated here.

\section{\label{sec:conclusion}Conclusion}

We have presented \autosindy, a hybrid discovery-then-solve framework that automates the feature-engineering bottleneck in the sparse identification of nonlinear dynamical systems. By using \pysr-based symbolic regression as a basis mining tool, coupling it with a principled collinearity-aware curation pipeline, and feeding the resulting custom library into \rev{the weak form of \sindy{}}, \autosindy{} recovers correct governing equations from noisy data without requiring any prior specification of the mathematical form. Benchmarks across \rev{six} canonical nonlinear systems\rev{, with noise of up to 20\% of each variable's standard deviation,} demonstrate consistent advantages in equation accuracy \rev{and} trajectory generalization\rev{, with equations close to the ground-truth complexity}. Unlike the large, black-box coefficient matrices produced by enriched \sindy{} libraries or the point-estimate expressions of standalone symbolic regression, the compact equations produced by \autosindy{} are immediately usable for mechanistic interpretation, control design, and hypothesis generation. Ultimately, this work establishes a pathway for automated scientific discovery in domains where the governing mathematics is genuinely unknown, including systems biology, neuroscience, climate modeling, and materials science. 

\section{Code and Data availability}
The complete source code, experimental configurations, and 
benchmark scripts are publicly available at 
\url{https://github.com/mabasiri95/AutoSINDy}.

\bibliography{references}

\appendix

\section{Effect of Noise on Raw Signal Quality}
\label{app:noise_levels}

\begin{figure*}[htbp]
  \centering
  \includegraphics[width=0.78\columnwidth]{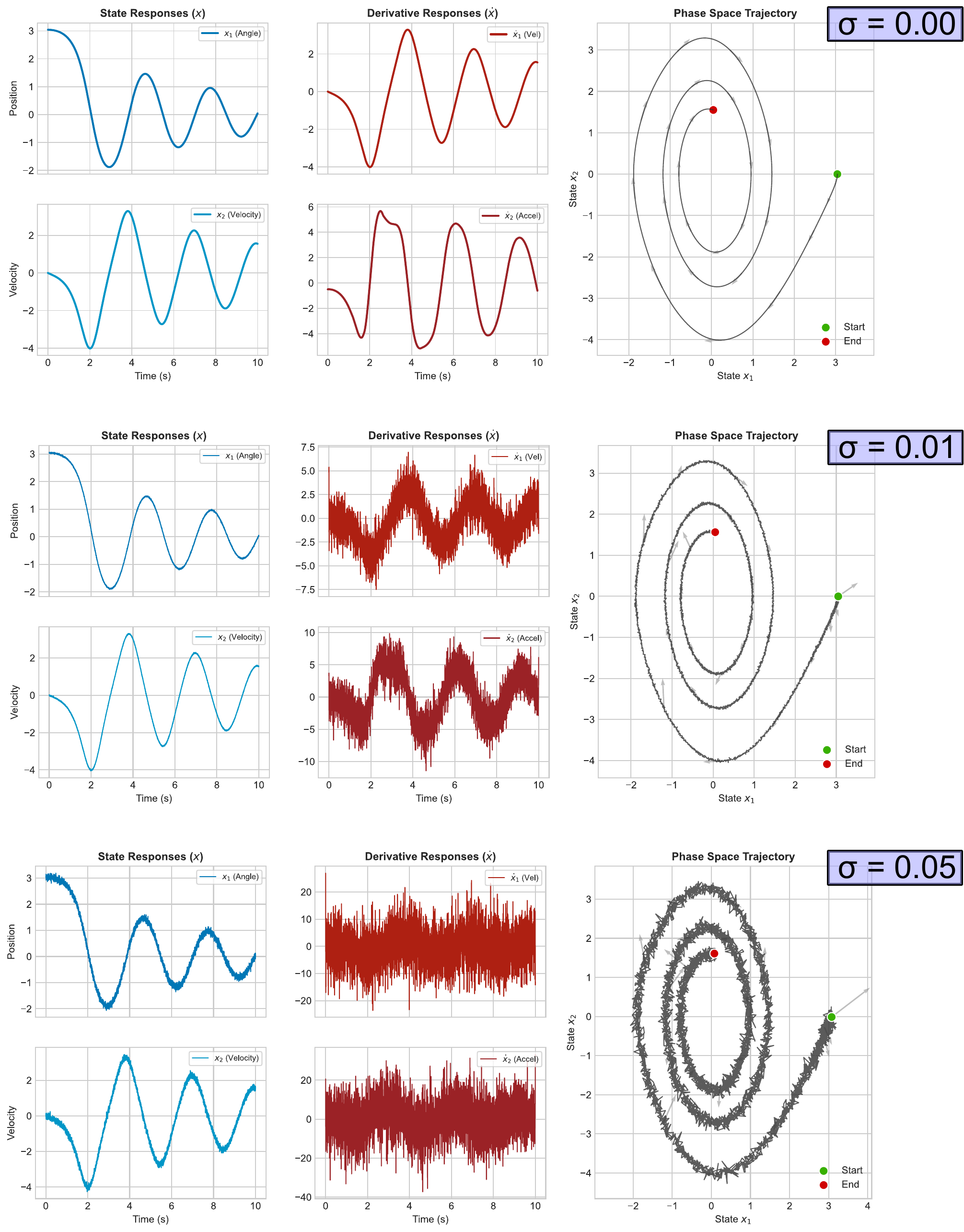}
\caption{Effect of measurement noise on system observability.
State responses ($x$, left column), numerical derivative estimates 
($\dot{x}$, center column), and phase-space trajectories (right column) 
for the damped pendulum at three noise levels: $\sigma = 0.00$ (top), 
$\sigma = 0.01$ (middle), and $\sigma = 0.05$ (bottom). While the 
state signals remain interpretable across all noise levels, the 
derivative estimates degrade rapidly, and the phase portrait becomes 
densely cluttered.}
\label{fig:app_noise}
\end{figure*}

This appendix supplements Sec.~\ref{subsec:data} of the main text.
Figure~\ref{fig:app_noise} illustrates how the noise level $\sigma$
affects the raw input data available to all methods.
The left columns show the state trajectories $x_1$ and $x_2$, which
remain visually smooth even at high noise because noise is injected
additively on the state.  The centre columns show the numerically
estimated time derivatives $\dot{x}_1$ and $\dot{x}_2$, which serve
as the regression targets for all \sindy-based methods and which
degrade sharply with noise: at $\sigma = 0.05$, the derivative signal
is almost entirely obscured.  The right column shows the corresponding
phase-space trajectory; the flow arrows become increasingly erratic
as noise grows, reflecting the corrupted derivative estimates.

{\color{revcolor}
The noise is added to the states. At each time point, each state variable
receives Gaussian noise with a standard deviation of $\sigma$ times the standard
deviation of that variable over the whole trajectory. Estimating the
derivatives amplifies this noise. Table~\ref{tab:derivative_noise} quantifies
the effect for all six systems. For each state variable, it compares the
derivatives estimated from the noisy states, as in the main study, with the
derivatives estimated in the same way from the noise-free states. It gives the
standard deviation of the difference as a percentage of the standard deviation
of the noise-free derivative, over the whole trajectory, as for $\sigma$. For
the five oscillators, the error is 28\% to 62\% at $\sigma = 0.01$, 139\% to
308\% at $\sigma = 0.05$ and 557\% to 1232\% at $\sigma = 0.20$, so from
$\sigma = 0.04$ on it is, on average, larger than the derivative signal itself.
For the Complex Lorenz system it is
much smaller (7\% to 140\%), because its states change much faster between two
samples than those of the oscillators.

}

\begin{table}[htbp]
\color{revcolor}
\centering
\caption{Error of the estimated derivatives caused by noise $\sigma$ on the states. The derivatives are estimated as in the main study. Each value is the standard deviation of the error as a percentage of the standard deviation of the noise-free derivative, over the whole trajectory. Values are averaged over the state variables and are medians over the 20 noise seeds of the main study.}
\label{tab:derivative_noise}
\renewcommand{\baselinestretch}{1}\small
\setlength{\tabcolsep}{7pt}
\begin{tabular}{@{}lccccccc@{}}
\toprule
 & \multicolumn{7}{c}{Noise $\sigma$ on the states} \\
\cmidrule(l){2-8}
\textbf{System} & 1\% & 2\% & 3\% & 4\% & 5\% & 10\% & 20\% \\
\midrule
Harmonic Oscillator & 45\% & 90\% & 135\% & 179\% & 224\% & 448\% & 897\% \\
Damped Pendulum & 62\% & 123\% & 185\% & 246\% & 308\% & 616\% & 1232\% \\
Modulated Oscillator & 43\% & 86\% & 129\% & 172\% & 215\% & 429\% & 859\% \\
Van der Pol & 28\% & 56\% & 83\% & 111\% & 139\% & 278\% & 557\% \\
Duffing Oscillator & 46\% & 93\% & 139\% & 185\% & 231\% & 463\% & 925\% \\
Complex Lorenz & 7\% & 14\% & 21\% & 28\% & 35\% & 70\% & 140\% \\
\bottomrule
\end{tabular}
\end{table}

{\color{revcolor}
\section{All methods on noise-free data}
\label{app:sindy_check}

This appendix supplements Sec.~\ref{subsec:baselines} of the main text.
It checks whether each method identifies the six systems from noise-free
data. Table~\ref{tab:clean_data} gives the $R^2$ on the training data and on
the new trajectory. We refitted the two \sindy{} baselines on these data with the
same training split as in the main study. For both, we replaced the
Polynomial--Fourier library of the main study by a polynomial library of
degree~3. For Standard \sindy{}, we also lowered the threshold from 0.21, the
value of the main study, to 0.15 and 0.1. All three values are below the
smallest true coefficient (0.25). The last five rows of
Table~\ref{tab:clean_data} are the noise-free runs of the main study.

Without noise and with the polynomial library, both \sindy{} baselines reach a
median new-trajectory $R^2$ above 0.99 on all six systems. With the
Polynomial--Fourier library, Standard \sindy{} does so on five systems and fails
only on the Duffing oscillator, at all three thresholds. Its lower recovery rates
in Table~\ref{tab:system_master_summary} therefore come from the noisy runs,
except on the Duffing oscillator. Weak \sindy{} fails on the harmonic, modulated
and Duffing oscillators, and Standard \pysr{} on the Van der Pol oscillator and
the Complex Lorenz system, where it misses true terms. \autosindy{} reaches a
median above 0.99 on all six systems.

The true terms of the Duffing oscillator are in the Polynomial--Fourier library,
but its training data cover only a narrow range of $x_0$
($0.69 \le x_0 \le 1.32$). On this range, many library terms are almost
combinations of others. The data can therefore hardly tell these terms apart, and both \sindy{} baselines return wrong equations for $\dot{x}_1$,
which fail on the new trajectory. On the harmonic oscillator, the model of Weak
\sindy{} does not even fit the training data. On the modulated oscillator, it
fits them but fails on the new trajectory. With the polynomial library, both
baselines find exactly the true terms of the Duffing oscillator, and so does
\autosindy{} in most of its noise-free runs of the main study.

With the polynomial library, both baselines also find exactly the true terms for the
harmonic and Van der Pol oscillators, whose equations are polynomial. For the
damped pendulum, the modulated oscillator and the Complex Lorenz system, whose
equations contain sines or cosines, they can only approximate the true equations
(Appendix~\ref{app:pendulum_degrees} examines the pendulum).

A small library that matches the form of the equations therefore works well,
but choosing it requires knowing that form. The main study asks how well each
method does without this knowledge, so its baselines use the
Polynomial--Fourier library: a fixed library close to the terms that
\autosindy's operators can build (polynomials, sines and cosines, and their
products).

}

\begin{table}[htbp]
\color{revcolor}
\centering
\caption{Derivative $R^2$ on noise-free data. Each cell gives the $R^2$ on the training data and on the new trajectory, as medians over 20 runs. The first two rows use a polynomial library of degree 3 in place of the Polynomial--Fourier library. The next two change the threshold of Standard SINDy, which is 0.21 in the main study. The last five rows use the settings of the main study.}
\label{tab:clean_data}
\renewcommand{\baselinestretch}{1}\footnotesize
\setlength{\tabcolsep}{3pt}
\begin{tabular}{@{}lcccccc@{}}
\toprule
 & Harmonic & Pendulum & Modulated & Van der Pol & Duffing & Complex Lorenz \\
\midrule
\multicolumn{7}{@{}l}{\textit{Polynomial library of degree 3}} \\
Standard SINDy & $1.000$/$1.000$ & $0.999$/$0.999$ & $1.000$/$1.000$ & $1.000$/$1.000$ & $1.000$/$1.000$ & $0.995$/$0.994$ \\
Weak SINDy & $1.000$/$1.000$ & $0.999$/$0.997$ & $1.000$/$1.000$ & $1.000$/$1.000$ & $1.000$/$1.000$ & $0.995$/$0.994$ \\
\midrule
\multicolumn{7}{@{}l}{\textit{Standard SINDy at other thresholds}} \\
Threshold 0.15 & $1.000$/$1.000$ & $0.995$/$0.994$ & $0.995$/$0.994$ & $1.000$/$1.000$ & $0.987$/$-1.165$ & $0.995$/$0.994$ \\
Threshold 0.1 & $1.000$/$1.000$ & $0.999$/$0.999$ & $0.995$/$0.994$ & $1.000$/$1.000$ & $0.987$/$-2.653$ & $0.995$/$0.994$ \\
\midrule
\multicolumn{7}{@{}l}{\textit{Settings of the main study}} \\
AutoSINDy & $1.000$/$1.000$ & $1.000$/$1.000$ & $1.000$/$1.000$ & $1.000$/$1.000$ & $1.000$/$1.000$ & $0.995$/$0.994$ \\
AutoSINDy-STLSQ & $1.000$/$1.000$ & $1.000$/$1.000$ & $1.000$/$1.000$ & $1.000$/$1.000$ & $1.000$/$1.000$ & $0.995$/$0.994$ \\
Standard SINDy & $1.000$/$1.000$ & $0.990$/$0.991$ & $0.995$/$0.994$ & $1.000$/$1.000$ & $0.499$/$0.473$ & $0.995$/$0.994$ \\
Weak SINDy & $0.091$/$-1.122$ & $0.998$/$0.992$ & $0.999$/$0.950$ & $0.995$/$0.995$ & $0.462$/$0.258$ & $0.995$/$0.994$ \\
Standard PySR & $1.000$/$1.000$ & $1.000$/$1.000$ & $1.000$/$1.000$ & $0.903$/$0.900$ & $1.000$/$1.000$ & $0.954$/$0.958$ \\
\bottomrule
\end{tabular}
\end{table}

{\color{revcolor}
\section{Damped pendulum with polynomial libraries of higher degree}
\label{app:pendulum_degrees}

A polynomial library can approximate $\sin x_0$, for example through its Taylor
series $x_0 - x_0^3/3! + x_0^5/5! - \dots$ This appendix tests how well such an
approximation works on the damped pendulum of the main study,
$\dot{x}_1 = -0.25\,x_1 - 5\sin x_0$, whose training data contain large swings
($-1.88 \le x_0 \le 2.70$). We fitted \sindy{} with polynomial libraries of
degree 3, 5 and 7 (10, 21 and 36 terms per equation) and the optimizer of the
main study, at three thresholds: 0.21, 0.01 and 0.001.
The data are those of the main study at $\sigma = 0$ and 0.01 (the first five
noise seeds each), and the \autosindy-STLSQ models are those of the main study on
the same data. \autosindy-STLSQ uses the same optimizer with the threshold 0.21,
so at this threshold the two differ only in the library. Besides the derivative $R^2$ on the new trajectory, we simulate each model
for 10 time units from two starts: the new trajectory's start,
$(\pi - 0.5, 0.2)$, and $(\pi - 0.1, 0)$, a larger swing than any in the
training data. Table~\ref{tab:pendulum_degrees} gives the results.

At 0.21, the threshold of the main study, all three polynomial models give the same cubic
model (four terms), which diverges when started from $(\pi - 0.1, 0)$.
\autosindy-STLSQ finds the true terms, $\dot{x}_1 = -5.01\sin x_0 - 0.252\,x_1$,
and simulates with $R^2 = 1.0000$ from both starts. At 0.01, degrees 5 and 7 add
an $x_0^5$ term (five terms) and match \autosindy-STLSQ on the new trajectory, but
fall behind it from $(\pi - 0.1, 0)$: $R^2 = 0.9693$ without noise and 0.9254 to
0.9734 at $\sigma = 0.01$. At 0.001, degree 7 overfits: without noise it keeps 36 terms
and fails on the new trajectory ($R^2 = 0.0209$).

Outside the training range, the polynomial models therefore fall behind
\autosindy-STLSQ at every threshold. At 0.21 they diverge, at 0.01 they need more
terms and reach a lower accuracy, and at 0.001 degree 7 overfits. \autosindy-STLSQ
finds the true law, with three terms, at the threshold of the main study.

}

\begin{table}[htbp]
\color{revcolor}
\centering
\caption{AutoSINDy-STLSQ and SINDy with polynomial libraries on the damped pendulum. The AutoSINDy-STLSQ models are those of the main study. The derivative $R^2$ is measured on the new trajectory. The simulation $R^2$ is measured over 10 time units from two starts. The first, $(\pi - 0.5, 0.2)$, is the start of the new trajectory. The second, $(\pi - 0.1, 0)$, gives a larger swing than any in the training data. Values are medians over the first five noise seeds of the main study. A simulation diverges when a state variable exceeds 100 times its true amplitude, and numbers such as (2/5) give how many of the five runs diverged.}
\label{tab:pendulum_degrees}
\renewcommand{\baselinestretch}{1}\small
\setlength{\tabcolsep}{7pt}
\begin{tabular}{@{}lccccc@{}}
\toprule
 & & & Derivative & \multicolumn{2}{c}{Simulation $R^2$ from} \\
\cmidrule(l){5-6}
Model & Threshold & Terms & $R^2$ & $(\pi - 0.5,\, 0.2)$ & $(\pi - 0.1,\, 0)$ \\
\midrule
\multicolumn{6}{@{}l}{\textit{Noise} $\sigma = 0$} \\
AutoSINDy-STLSQ & 0.21 & 3 & $1.0000$ & $1.0000$ & $1.0000$ \\
\addlinespace
\multirow{3}{*}{SINDy, polynomial degree 3} & 0.21 & 4 & $0.9990$ & $0.9831$ & diverges \\
 & 0.01 & 7 & $0.9994$ & $0.9974$ & diverges \\
 & 0.001 & 11 & $0.9995$ & $0.9988$ & diverges \\
\addlinespace
\multirow{3}{*}{SINDy, polynomial degree 5} & 0.21 & 4 & $0.9990$ & $0.9831$ & diverges \\
 & 0.01 & 5 & $1.0000$ & $1.0000$ & $0.9693$ \\
 & 0.001 & 11 & $1.0000$ & $0.9996$ & $0.9662$ \\
\addlinespace
\multirow{3}{*}{SINDy, polynomial degree 7} & 0.21 & 4 & $0.9990$ & $0.9831$ & diverges \\
 & 0.01 & 5 & $1.0000$ & $1.0000$ & $0.9693$ \\
 & 0.001 & 36 & $0.0209$ & diverges & diverges \\
\midrule
\multicolumn{6}{@{}l}{\textit{Noise} $\sigma = 0.01$} \\
AutoSINDy-STLSQ & 0.21 & 3 & $1.0000$ & $1.0000$ & $1.0000$ \\
\addlinespace
\multirow{3}{*}{SINDy, polynomial degree 3} & 0.21 & 4 & $0.9990$ & $0.9847$ & diverges \\
 & 0.01 & 7 & $0.9994$ & $0.9976$ & diverges \\
 & 0.001 & 17 & $0.9995$ & $0.9962$ & diverges \\
\addlinespace
\multirow{3}{*}{SINDy, polynomial degree 5} & 0.21 & 4 & $0.9990$ & $0.9812$ & diverges \\
 & 0.01 & 5 & $1.0000$ & $0.9998$ & $0.9734$ \\
 & 0.001 & 31 & $0.9990$ & $0.9889$ & $-0.0727$ (1/5) \\
\addlinespace
\multirow{3}{*}{SINDy, polynomial degree 7} & 0.21 & 4 & $0.9990$ & $0.9847$ & diverges \\
 & 0.01 & 5 & $1.0000$ & $0.9998$ & $0.9254$ \\
 & 0.001 & 67 & $0.9399$ & $0.7915$ (1/5) & diverges (3/5) \\
\bottomrule
\end{tabular}
\end{table}

{\color{revcolor}
\section{Sensitivity to the settings of \autosindy{}}
\label{app:settings}

This appendix supplements Sec.~\ref{subsec:hyperparams} of the main text.
We changed one setting of \autosindy{} at a time and kept all others as in
the main study. The tests use three systems (damped pendulum, Van der Pol
oscillator and Complex Lorenz system), $\sigma = 0.01$ and 0.05, and the 20
seeds of the main study, so each result is over 60 runs.
Table~\ref{tab:sensitivity} gives the results.

With chunks of 200 or 800 instead of 400 points, and with 2 or 5 instead of
3 chunks, the mean TPR changed by at most 0.04. Fewer chunks lowered the
share of runs with $R^2$ above 0.99 at $\sigma = 0.05$, from 97\% to 88\%
with 2 chunks. Five chunks raised it to 100\%, but made a run almost twice
as slow (46~s instead of 26~s) and the equations longer (median complexity
11.5 instead of 8). The results therefore change little with the chunks in
this range.

We also trained all five methods on the middle 50\%, 65\%, 80\% (default)
and 95\% of the same recording, so that each window contains the smaller
ones. This test uses the damped pendulum, the Van der Pol oscillator and the
Duffing oscillator, with the same noise levels and seeds.
Table~\ref{tab:windows} gives the results.

On the damped pendulum and the Van der Pol oscillator, the comparison does
not change with the amount of data. \autosindy{} keeps a median $R^2$ of at
least 0.99 with every window, and each baseline is either less accurate or
uses longer equations. The baselines gain most from more data. On the
pendulum at $\sigma = 0.01$, the median $R^2$ of Standard \sindy{} rises from
below zero with the 50\% window to 0.995 with the 80\% window. On the Duffing
oscillator, the result depends on whether the window includes the first large
swing, as discussed in Sec.~\ref{subsec:duffing_discussion}.

}

\begin{table}[htbp]
\color{revcolor}
\centering
\caption{Results of AutoSINDy with one setting changed at a time from the default, on the damped pendulum, the Van der Pol oscillator and the Complex Lorenz system (20 runs each, 60 runs per cell). $R^2$ is the derivative $R^2$ on the new trajectory. The complexity is the median at $\sigma = 0.05$ (at $\sigma = 0.01$ it is 8.0 for every setting), and the time is the median run time in seconds, measured with three runs in parallel. For every setting, the median $R^2$ is above 0.99, and no simulation failed.}
\label{tab:sensitivity}
\renewcommand{\baselinestretch}{1}\small
\begin{tabular}{@{}lcccccc@{}}
\toprule
 & \multicolumn{2}{c}{\begin{tabular}{@{}c@{}}Runs with\\ $R^2 > 0.99$ (\%)\end{tabular}} & \multicolumn{2}{c}{Mean TPR} & & \\
\cmidrule(lr){2-3} \cmidrule(lr){4-5}
Setting & $\sigma = 0.01$ & $\sigma = 0.05$ & $\sigma = 0.01$ & $\sigma = 0.05$ & Complexity & Time (s) \\
\midrule
Default (3 chunks of 400 points) & 98 & 97 & 0.90 & 0.83 & 8.0 & 26 \\
Chunks of 200 points & 98 & 93 & 0.90 & 0.81 & 8.0 & 25 \\
Chunks of 800 points & 98 & 93 & 0.92 & 0.81 & 8.0 & 38 \\
2 chunks & 95 & 88 & 0.88 & 0.79 & 8.0 & 19 \\
5 chunks & 100 & 100 & 0.92 & 0.81 & 11.5 & 46 \\
\bottomrule
\end{tabular}
\end{table}

\begin{table}[htbp]
\color{revcolor}
\centering
\caption{Results of all methods trained on the middle 50, 65, 80 or 95\% of the same recording (2500, 3250, 4000 or 4750 training points). The 80\% window ($^\dagger$) is the default used in all other results. Each cell gives the median derivative $R^2$ on the new trajectory and the median complexity over 20 runs. The entry $<0$ marks a median $R^2$ below zero, and a complexity of 0 means that every term was discarded. Bold marks the highest $R^2$ for each system and column and, among equal values, the lowest complexity.}
\label{tab:windows}
\renewcommand{\baselinestretch}{1}\small
\setlength{\tabcolsep}{4pt}
\begin{tabular}{@{}llcccc@{}}
\toprule
System & Model & 50\% & 65\% & 80\%$^\dagger$ & 95\% \\
\midrule
\multicolumn{6}{@{}l}{\textit{Noise} $\sigma = 0.01$} \\
\multirow{5}{*}{Damped Pendulum} & AutoSINDy & \textbf{0.991\,/\,3.0} & \textbf{1.000\,/\,5.0} & \textbf{1.000\,/\,5.0} & \textbf{1.000\,/\,5.0} \\
 & AutoSINDy-STLSQ & 0.991\,/\,3.5 & \textbf{1.000\,/\,5.0} & \textbf{1.000\,/\,5.0} & \textbf{1.000\,/\,5.0} \\
 & Standard SINDy & $<0$\,/\,81.0 & 0.962\,/\,30.5 & 0.995\,/\,22.0 & 0.994\,/\,16.0 \\
 & Weak SINDy & $<0$\,/\,205.5 & 0.911\,/\,199.5 & 0.982\,/\,224.0 & 0.997\,/\,217.5 \\
 & Standard PySR & 0.866\,/\,5.0 & 0.856\,/\,5.0 & 0.991\,/\,6.0 & 0.991\,/\,6.0 \\
\addlinespace
\multirow{5}{*}{Van der Pol} & AutoSINDy & \textbf{1.000\,/\,8.0} & \textbf{1.000\,/\,8.0} & \textbf{1.000\,/\,8.0} & \textbf{1.000\,/\,8.0} \\
 & AutoSINDy-STLSQ & 1.000\,/\,12.0 & \textbf{1.000\,/\,8.0} & 1.000\,/\,10.0 & \textbf{1.000\,/\,8.0} \\
 & Standard SINDy & 1.000\,/\,98.5 & 0.997\,/\,103.5 & 1.000\,/\,83.0 & 1.000\,/\,49.5 \\
 & Weak SINDy & 0.999\,/\,169.0 & 0.982\,/\,91.0 & 0.998\,/\,172.5 & 0.978\,/\,211.0 \\
 & Standard PySR & 0.901\,/\,9.0 & 0.899\,/\,9.0 & 0.901\,/\,9.0 & 0.899\,/\,9.0 \\
\addlinespace
\multirow{5}{*}{Duffing Osc.} & AutoSINDy & $<0$\,/\,5.0 & 0.823\,/\,4.5 & \textbf{1.000\,/\,7.0} & \textbf{1.000\,/\,7.0} \\
 & AutoSINDy-STLSQ & $<0$\,/\,6.0 & \textbf{0.834\,/\,6.0} & \textbf{1.000\,/\,7.0} & \textbf{1.000\,/\,7.0} \\
 & Standard SINDy & $<0$\,/\,9.0 & $<0$\,/\,9.0 & $<0$\,/\,7.0 & $<0$\,/\,11.0 \\
 & Weak SINDy & $<0$\,/\,0.0 & $<0$\,/\,0.0 & $<0$\,/\,0.0 & $<0$\,/\,4.0 \\
 & Standard PySR & $<0$\,/\,4.5 & $<0$\,/\,5.0 & 0.904\,/\,9.5 & 0.947\,/\,5.0 \\
\midrule
\multicolumn{6}{@{}l}{\textit{Noise} $\sigma = 0.05$} \\
\multirow{5}{*}{Damped Pendulum} & AutoSINDy & \textbf{0.993\,/\,5.0} & \textbf{0.991\,/\,5.0} & \textbf{1.000\,/\,5.0} & \textbf{0.995\,/\,6.0} \\
 & AutoSINDy-STLSQ & 0.979\,/\,16.5 & 0.981\,/\,14.5 & 0.995\,/\,9.5 & 0.994\,/\,6.0 \\
 & Standard SINDy & $<0$\,/\,399.0 & $<0$\,/\,349.5 & 0.898\,/\,314.5 & 0.975\,/\,337.5 \\
 & Weak SINDy & $<0$\,/\,196.0 & 0.804\,/\,215.0 & 0.960\,/\,250.0 & 0.995\,/\,231.0 \\
 & Standard PySR & 0.855\,/\,5.0 & 0.850\,/\,5.0 & 0.890\,/\,5.0 & 0.863\,/\,5.0 \\
\addlinespace
\multirow{5}{*}{Van der Pol} & AutoSINDy & \textbf{1.000\,/\,8.0} & \textbf{1.000\,/\,8.0} & \textbf{1.000\,/\,8.0} & \textbf{1.000\,/\,8.0} \\
 & AutoSINDy-STLSQ & 0.999\,/\,18.5 & 0.999\,/\,17.0 & 0.999\,/\,12.0 & 0.999\,/\,14.5 \\
 & Standard SINDy & 0.991\,/\,327.0 & 0.991\,/\,337.0 & 0.992\,/\,333.0 & 0.993\,/\,334.5 \\
 & Weak SINDy & 0.996\,/\,170.5 & 0.957\,/\,121.5 & 0.993\,/\,176.0 & 0.964\,/\,177.0 \\
 & Standard PySR & 0.799\,/\,8.0 & 0.800\,/\,8.0 & 0.801\,/\,8.0 & 0.801\,/\,8.0 \\
\addlinespace
\multirow{5}{*}{Duffing Osc.} & AutoSINDy & $<0$\,/\,6.0 & $<0$\,/\,7.0 & \textbf{0.999\,/\,7.5} & \textbf{0.967\,/\,7.0} \\
 & AutoSINDy-STLSQ & $<0$\,/\,7.5 & $<0$\,/\,7.5 & 0.963\,/\,8.0 & 0.923\,/\,7.0 \\
 & Standard SINDy & $<0$\,/\,17.5 & $<0$\,/\,16.5 & $<0$\,/\,13.5 & $<0$\,/\,19.0 \\
 & Weak SINDy & $<0$\,/\,0.0 & $<0$\,/\,0.0 & $<0$\,/\,0.0 & $<0$\,/\,0.0 \\
 & Standard PySR & $<0$\,/\,4.0 & $<0$\,/\,4.5 & $<0$\,/\,5.0 & $<0$\,/\,5.5 \\
\bottomrule
\end{tabular}
\end{table}

\section{Per-System Identification Results}
\label{app:per_system_results}

\begin{figure*}[htbp]
  \centering
  \includegraphics[width=\textwidth]{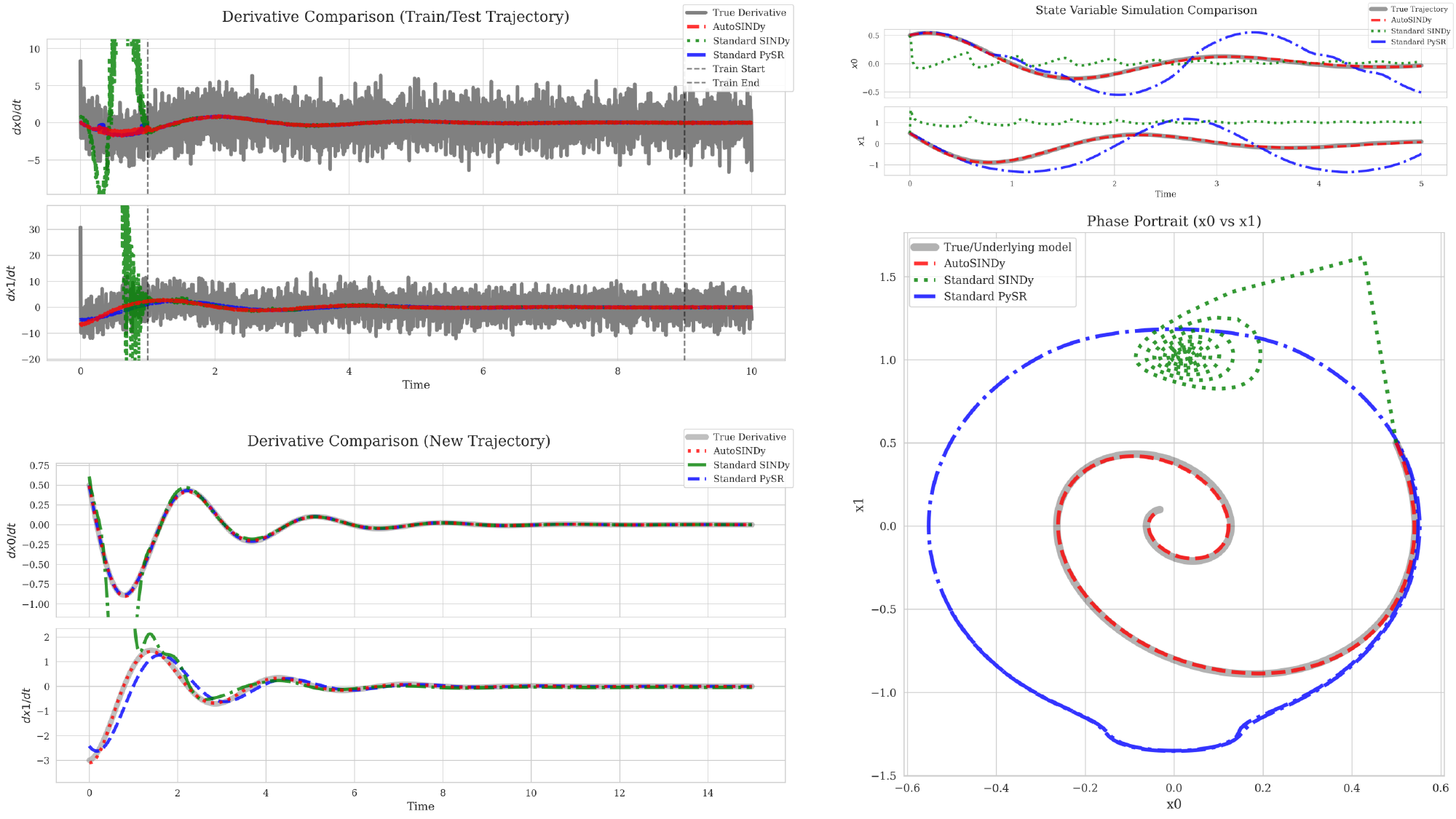}
  \caption{%
    Identification results for the \textbf{harmonic oscillator}
    ($\dot{x}_0 = x_1$, $\dot{x}_1 = -k_1 x_0 - k_2 x_1$) at $\sigma = 0.08$.
    \autosindy{}\rev{-STLSQ} recovers the true dynamics and \rev{reproduces
    the inward-spiralling phase portrait}.  Standard \sindy{} fails to generalize to the data beyond the training window boundaries and
    exhibits trajectory drift due to library bloat. Also, standard \pysr{} produces a limit-cycle-like
    divergence in the phase portrait, indicating a missed damping term.%
  }
  \label{fig:app_harmonic}
\end{figure*}

\begin{figure*}[htbp]
  \centering
  \includegraphics[width=\textwidth]{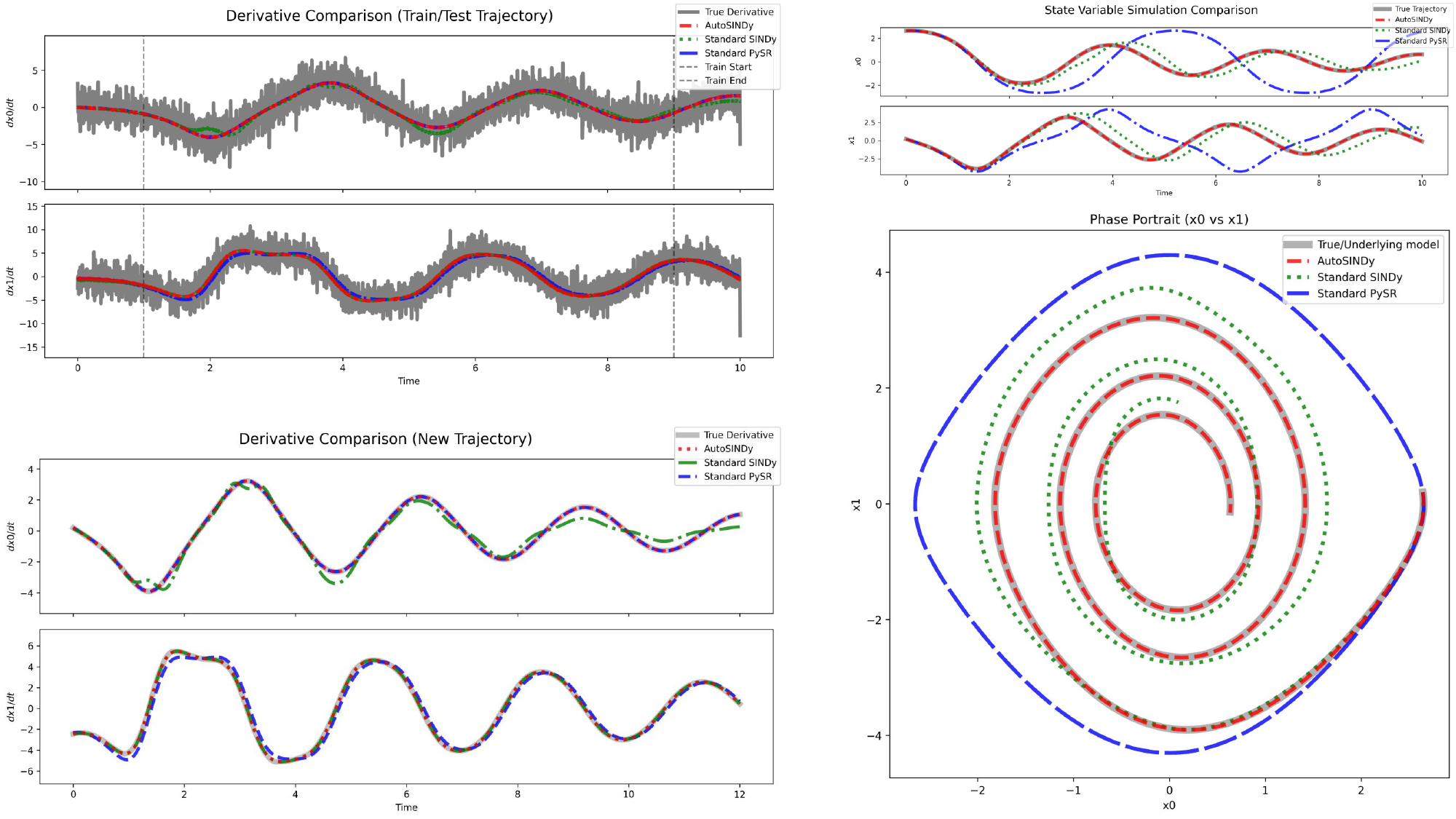}
  \caption{%
    Identification results for the \textbf{damped pendulum}
    ($\dot{x}_0 = x_1$, $\dot{x}_1 = -bx_1 - c\sin(x_0)$\rev{)} at $\sigma = 0.01$.
    \autosindy{}\rev{-STLSQ} correctly recovers the nonlinearity
    and reproduces the inward-spiralling phase portrait with near-zero simulation error.
    Standard \sindy{} retains numerous spurious 
  polynomial--Fourier cross-terms, degrading
  simulation fidelity.
    Standard \pysr{} misidentifies the coefficient of the damping
    term, causing the trajectory not to decay.%
  }
  \label{fig:app_pendulum}
\end{figure*}

\begin{figure*}[htbp]
  \centering
  \includegraphics[width=\textwidth]{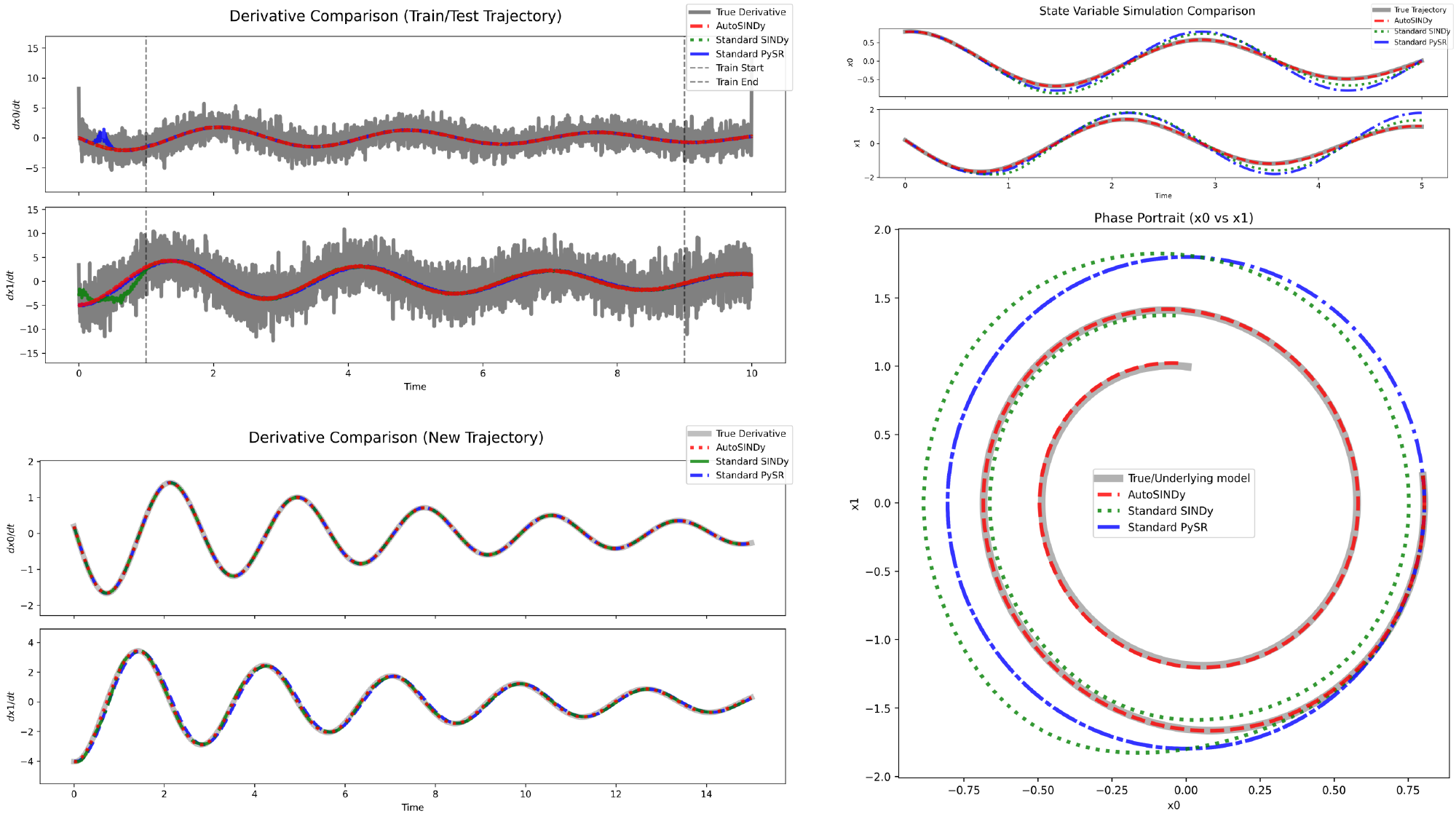}
  \caption{%
    Identification results for the \textbf{modulated oscillator}
    ($\dot{x}_0 = x_1$,
    $\dot{x}_1 = -bx_1\cos(x_0) - kx_0$) at $\sigma = 0.03$.
    \autosindy{}\rev{-STLSQ} recovers the damping behavior
    and produces a stable decaying-oscillation trajectory.
    Both baselines fail to identify the exact dynamics, resulting in errors in both the derivative comparison beyond the training data and in the phase portrait trajectories.%
  }
  \label{fig:app_modulated}
\end{figure*}

\begin{figure*}[htbp]
  \centering
  \includegraphics[width=\textwidth]{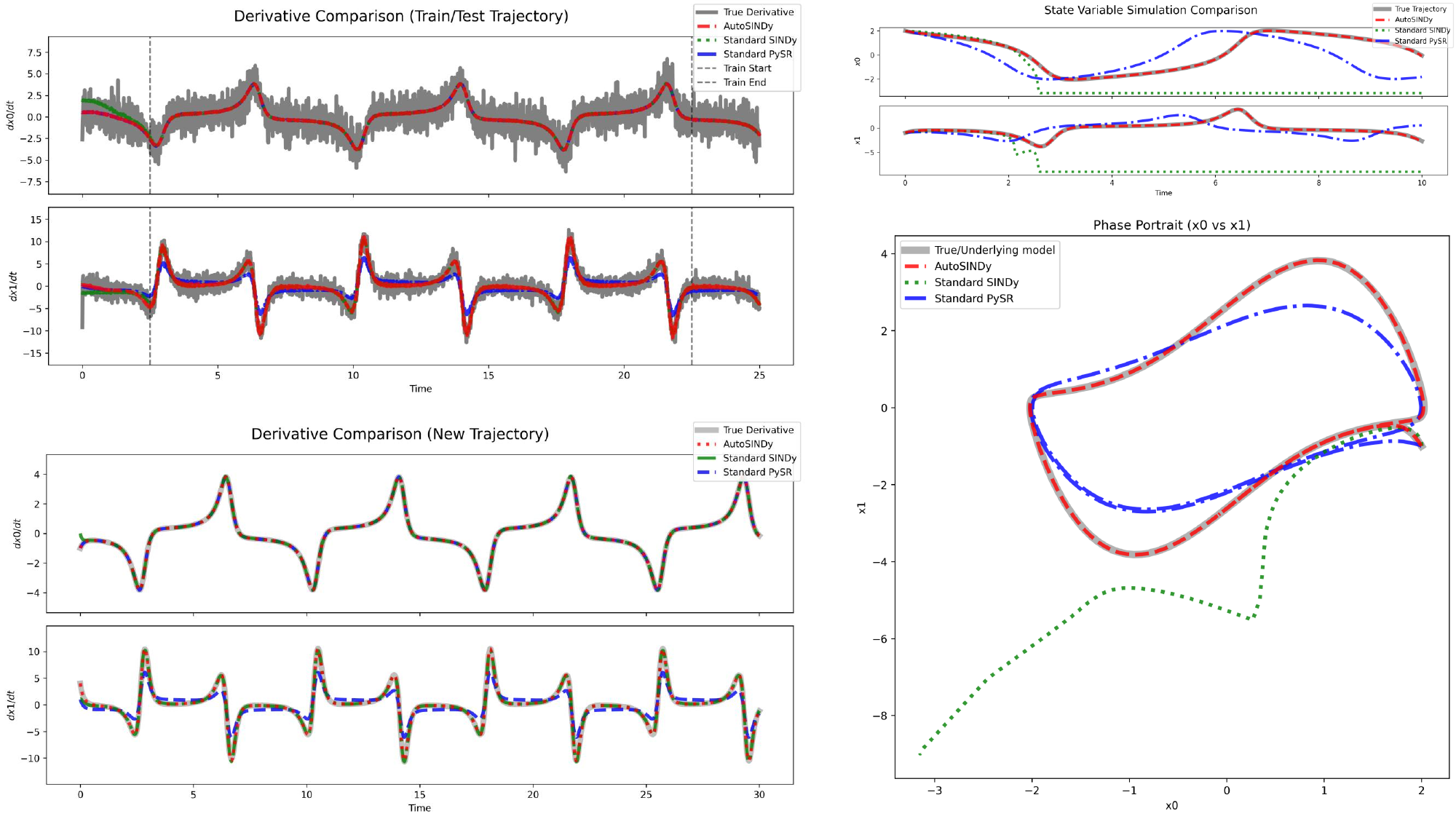}
  \caption{%
    Identification results for the \textbf{Van der Pol oscillator}
    ($\dot{x}_0 = x_1$,
    $\dot{x}_1 = \mu(1 - x_0^2)x_1 - x_0$) at $\sigma = 0.02$.
    \autosindy{}\rev{-STLSQ} recovers the limit-cycle structure and the
    correct dynamics, yielding
    a stable phase portrait.  Standard \sindy{}'s bloated
    equations \rev{cause} trajectory divergence.  Standard \pysr{} \rev{misestimates
    the dynamics and settles on a different limit cycle}.%
  }
  \label{fig:app_vanderpol}
\end{figure*}

\begin{figure*}[htbp]
  \centering
  \includegraphics[width=\textwidth]{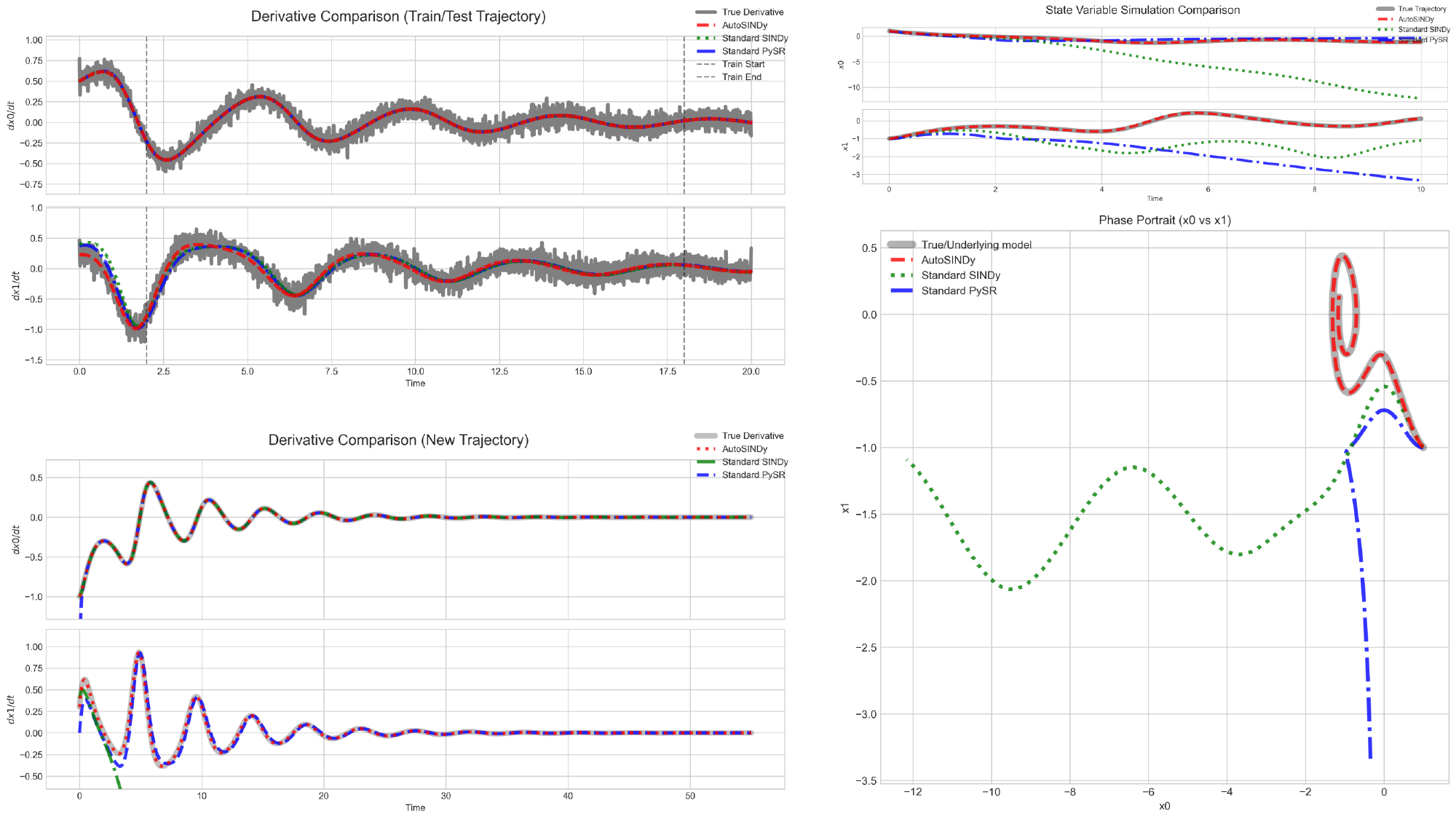}
  \caption{%
    Identification results for the \textbf{Duffing oscillator}
    ($\dot{x}_0 = x_1$,
    $\dot{x}_1 = -\delta x_1 - \alpha x_0 - \beta x_0^3$) at $\sigma = 0.05$.
    This is the most challenging system for all methods due to the
    bistable potential and high derivative sensitivity near the
    equilibria.  \autosindy{}\rev{-STLSQ}
     recovered the true dynamics and
    the phase portrait closely matches the true trajectory.%
  }
  \label{fig:app_duffing}
\end{figure*}

\begin{figure*}[htbp]
  \centering
  \includegraphics[width=\textwidth]{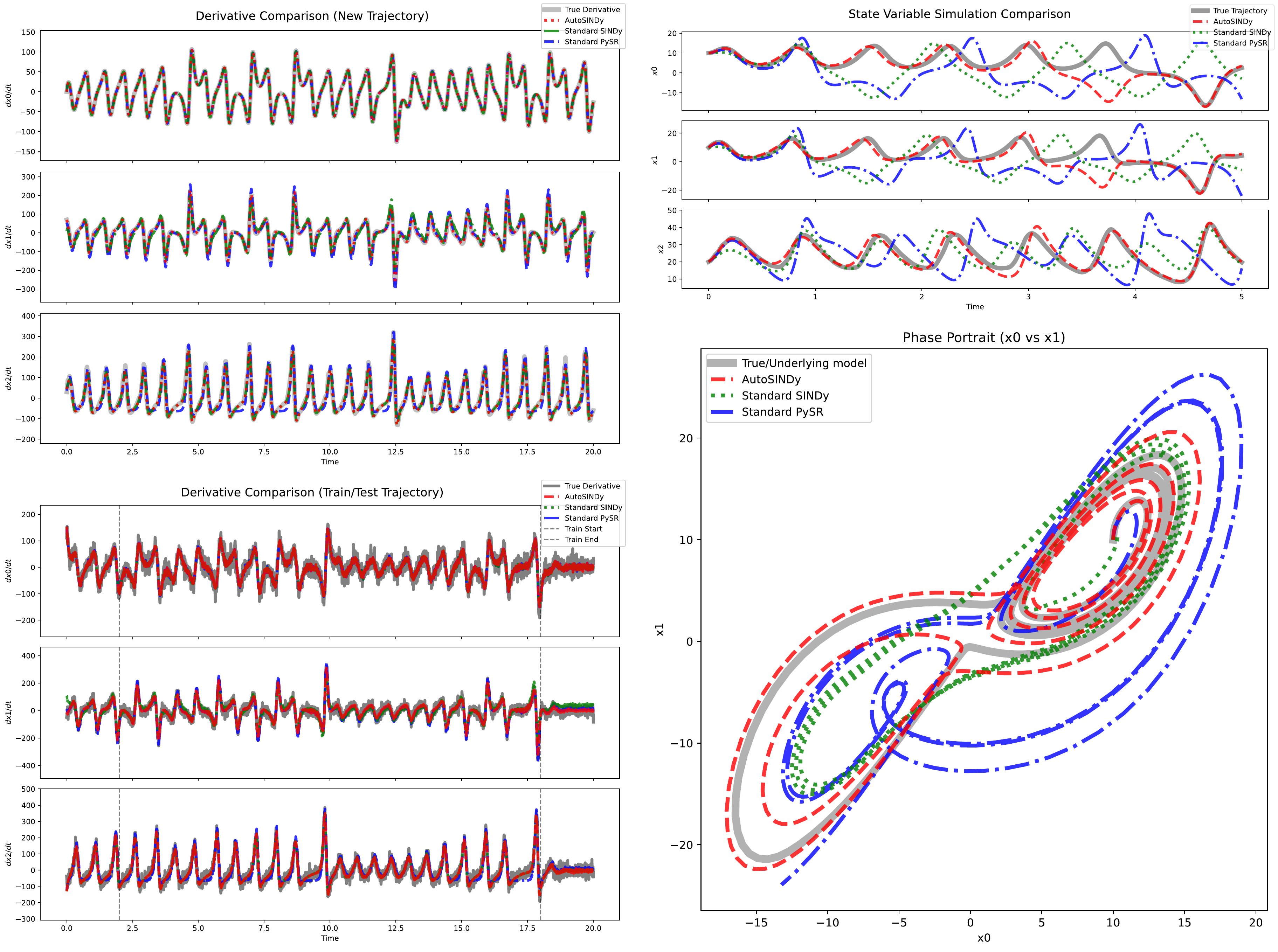}
  \caption{%
    Identification results for the \textbf{Complex Lorenz system}
    ($\dot{x}_0 = \sigma(x_1-x_0)$,
    $\dot{x}_1 = x_0(\rho - x_2) - x_1$,
    $\dot{x}_2 = x_0 x_1 - \beta x_2 + \gamma x_1\sin(x_0+x_2)$) at $\sigma = 0.01$.
    The three-dimensional state space is visualized via the $x_0$--$x_1$
    phase portrait.  \autosindy{}\rev{-STLSQ} closely tracks parts of the true
    trajectory using a simple discovered equation.%
  }
  \label{fig:app_lorenz}
\end{figure*}

This appendix supplements Sec.~\ref{subsec:results_noise} of the main
text.  Figures~\ref{fig:app_harmonic}--\ref{fig:app_lorenz} show
representative single-trial identification results for each of the six
benchmark systems. For each system, 
we show: (i) derivative fits on the training/test trajectory, 
(ii) generalization to an unseen trajectory, (iii) state-variable 
simulation comparison, and (iv) phase portrait. Each figure follows
the same layout. The left panels show derivative comparisons on the
training/test trajectory\rev{, where} vertical
dashed lines mark the training window boundaries (upper left), and an independent out-of-sample unseen
trajectory (lower left). The upper right panel shows forward-simulated
state variables from a held-out 
initial condition over time compared against the ground truth. The lower right
panel shows the phase portrait comparing 
the predicted dynamics against the true 
underlying model.  \autosindy{}\rev{-STLSQ} (red dashed\rev{, labelled
\autosindy{} in the figures}),
Standard \sindy{} (green dotted), and Standard \pysr{} (blue
dash-dot) are compared against the true underlying model (gray solid).

\end{document}